\documentclass{article}

\usepackage[final]{neurips_2021}
\usepackage{bbm}
\usepackage{amssymb}
\usepackage{amsmath}
\usepackage[normalem]{ulem}
\usepackage{bm}
\usepackage{graphicx}
\usepackage{multirow}
\usepackage{caption}
\usepackage{paralist}
\usepackage{subcaption}
\usepackage{diagbox}
\usepackage{pifont}
\usepackage{mathtools}

\usepackage[most]{tcolorbox}

\usepackage[utf8]{inputenc}
\usepackage[T1]{fontenc}
\usepackage{hyperref}
\hypersetup{
    colorlinks=true,
    linkcolor=blue,
    filecolor=magenta,      
    urlcolor=cyan,
    citecolor=blue,
}
\usepackage{url}
\usepackage{booktabs}
\usepackage{amsfonts}
\usepackage{nicefrac}
\usepackage{microtype}
\usepackage{xcolor}
\usepackage{enumitem}

\renewcommand{\eqref}[1]{Equation~\ref{#1}}

\newcommand{\secref}[1]{Section~\ref{#1}}

\newcommand{\D}{{\mathcal D}}

\newcommand{\x}{{\mathbf x}}

\newcommand{\btheta}{{\boldsymbol \theta}}

\newcommand{\cmark}{\ding{51}}%
\newcommand{\xmark}{\ding{55}}%

\graphicspath{{figures/}}

\title{Disentangling the Roles of Curation, Data-Augmentation and the Prior in the \\ Cold Posterior Effect}

\author{
  Lorenzo Noci$^*$\\
  Dept of Computer Science \\
  ETH Z\"urich\\
  \texttt{\small lorenzo.noci@inf.ethz.ch} \\
  \And
  Kevin Roth$^*$\\
  Dept of Computer Science \\
  ETH Z\"urich\\
  \texttt{\small kevin.roth@inf.ethz.ch} \\
  \And
  Gregor Bachmann$^*$\\
  Dept of Computer Science \\
  ETH Z\"urich\\
  \texttt{\small gregor.bachmann@inf.ethz.ch} \\
  \And
  Sebastian Nowozin\\
  Microsoft Research \\
  Cambridge, UK \\
  \texttt{\small Sebastian.Nowozin@microsoft.com} \\
  \And
  Thomas Hofmann \\
  Dept of Computer Science \\
  ETH Z\"urich\\
  \texttt{\small thomas.hofmann@inf.ethz.ch} \\   
}

\begin{document}

\maketitle

\begin{abstract}

The ``\emph{cold posterior effect}'' (CPE) in Bayesian deep learning describes the uncomforting observation that the predictive performance of Bayesian neural networks can be significantly improved if the Bayes posterior is artificially sharpened using a temperature parameter $T < 1$.
The CPE is problematic in theory and practice and since the effect was identified many researchers have proposed hypotheses to explain the phenomenon. However, despite this intensive research effort the effect remains poorly understood.
In this work we provide novel and nuanced evidence relevant to existing explanations for the cold posterior effect, disentangling three hypotheses:
1. The \emph{dataset curation hypothesis} of~\cite{aitchison2020statistical}:
we show empirically that the CPE does not arise in a real curated data set but can be produced in a controlled experiment with varying curation strength.
2. The \emph{data augmentation hypothesis} of~\cite{izmailov2021bayesian} and~\cite{fortuin2021bayesian}:
we show empirically that data augmentation is sufficient but not necessary for the CPE to be present.
3. The \emph{bad prior hypothesis} of~\cite{wenzel2020good}:
we use a simple experiment evaluating the relative importance of the prior and the likelihood, strongly linking the CPE to the prior.
Our results demonstrate how the CPE can arise in isolation from synthetic curation, data augmentation, and bad priors.
Cold posteriors observed ``in the wild'' are therefore unlikely to arise from a single simple cause; as a result, we do not expect a simple ``fix'' for cold posteriors. 
\end{abstract}


\section{Introduction}

Deep neural networks have achieved great success in predictive accuracy for supervised learning tasks.
Unfortunately, however, they still fall short in giving useful estimates of their predictive \textit{uncertainty}, i.e.\ meaningful confidence values for how certain the model is about its predictions \citep{ovadia2019modeluncertainty}.
Quantifying uncertainty is especially crucial in real-world settings, which often involve data distributions that are shifted from the one seen during training \citep{quionero2009dataset}.

Bayesian deep learning combines deep learning with Bayesian probability theory.
\textit{Bayesian neural networks} (BNNs) learn a distribution over model parameters or equivalently sample an ensemble of likely models given the data, 
promising better generalization performance and principled uncertainty quantification (robust predictions) \citep{neal1995bayesianneuralnetworks, mackay1992practical, dayan1995helmholtz}.

In Bayesian deep learning we either learn a distribution $q(\btheta)$ over models compatible with the data, i.e.\ $q(\btheta)\simeq p(\btheta|\D)$, or we \textit{sample} an ensemble of models 
$\btheta_{1}, \dots, \btheta_{K} \sim p(\btheta | \D)$ 
from the posterior over likely models 
\begin{equation}
p(\btheta|\mathcal{D}) \propto p(\mathcal{D}|\btheta) \, p(\btheta),
\end{equation}
where in the i.i.d.\ setting $p(\mathcal{D}|\btheta) = \prod_{i=1}^n p(y_i|\x_i,\btheta)$ is the likelihood, relating the model we want to learn to the observations $\D = \{(\x_i,y_i)\}_{i=1}^n$,  
and $p(\btheta)$ is a proper prior, e.g.\ a Gaussian density.

BNN predictions involve model averaging:
rather than betting everything on a single point estimate of the parameters, 
we predict on a new instance $\x$ by \textit{averaging} over all likely models compatible with the data, 
\vspace{-2mm}
\begin{equation}
    p(y|\x,\mathcal{D}) = \int p(y|\x,\btheta) \, p(\btheta|\mathcal{D}) \,\textrm{d}\btheta \,\,\,\ 
    \simeq \,\ \sum_{k=1}^K p(y | \x, \btheta_{k}) \, ,\,\ \textnormal{where}\,\ \btheta_{k} \sim p(\btheta | \D) \, .
    \label{eqn:intro:postpred}
\end{equation}
\eqref{eqn:intro:postpred} is also known as the \emph{posterior predictive} or \emph{Bayesian model average}.
Note that, in practice, solving the integral exactly is impossible.
However, we can approximate it via Monte Carlo sampling using an ensemble of models $\btheta_{k} \sim p(\btheta | \D)$, see \secref{eq:BNNposteriorinference} in the Appendix for further details.

The two main inference paradigms to learn a distribution over model parameters $q(\btheta)\simeq p(\btheta|\D)$ respectively to sample from the posterior $\btheta_{1}, \dots, \btheta_{K} \sim p(\btheta | \D)$ are 
Variational Bayes (VB) \citep{hinton1993mdl, mackay1995ensemblelearning, barber1998ensemblelearning, blundell2015weightuncertainty} and Markov Chain Monte Carlo (MCMC) \citep{neal1995bayesianneuralnetworks, welling2011bayesian, chen2014stochastichmc, ma2015complete}. 
We will focus on MCMC methods as they are simple to implement and can be scaled to large models and datasets 
when used with stochastic minibatch gradients (SG-MCMC) \citep{welling2011bayesian, chen2014stochastichmc, li2016preconditionedsgld}.

In recent years, the Bayesian deep learning community has developed increasingly accurate and efficient approximate inference procedures for deep BNNs (cf.\ references in the paragraph above).
Despite this algorithmic progress, however, important questions surrounding BNNs remain unanswered to this day.
A recent and particularly prominent one concerns the ``\emph{cold posterior effect}'' (CPE), which describes the observation that the predictive performance of BNNs can be significantly improved if the Bayes posterior is artificially sharpened $p(\btheta|\mathcal{D})^{1/T}$ using a temperature parameter $T < 1$ \citep{wenzel2020good}.
Such cold posteriors sharply deviate from the Bayesian paradigm but are commonly used as heuristics in practice, see Section~2.3 in \citep{wenzel2020good}.

The CPE is problematic in theory and practice, and since the effect was identified many researchers have proposed hypotheses to explain the phenomenon.
There has been an ongoing debate questioning the roles of isotropic Gaussian priors \citep{wenzel2020good, zeno2020cold, fortuin2021bayesian}, the likelihood model \citep{aitchison2020statistical}, inaccurate inference \citep{izmailov2021bayesian, wenzel2020good}, and data augmentation \citep{izmailov2021bayesian, fortuin2021bayesian}. 
However, despite this intensive research effort the effect remains poorly understood.

\paragraph{Contributions:}
We provide novel and nuanced evidence relevant to existing explanations for the cold posterior effect, 
disentangling the roles of curation, data augmentation and the prior:

\begin{itemize}[leftmargin=8mm]
\item The \emph{dataset curation hypothesis} of~\cite{aitchison2020statistical}:
we show empirically that the CPE does not arise in a real curated data set but can be produced in a controlled experiment with varying curation strength.
\item The \emph{data augmentation hypothesis} of~\cite{izmailov2021bayesian} and~\cite{fortuin2021bayesian}:
we show empirically that data augmentation is sufficient but not necessary for the CPE to be present.
\item The \emph{bad prior hypothesis} of~\cite{wenzel2020good}:
we use a simple experiment evaluating the relative importance of the prior and the likelihood, strongly linking the CPE to the prior.
\end{itemize}

Our results demonstrate how the CPE can arise in isolation from synthetic curation, data augmentation, and bad priors.
In fact, we are able to reproduce the cold posterior effect with each of the three factors alone.
Cold posteriors observed ``in the wild'' are therefore unlikely to arise from a single cause; 
as a result, we do not expect a simple ``fix'' for cold posteriors.


\section{Cold Posteriors: Background \& Related Work}
\label{sec:coldposteriors}

The ``\emph{cold posterior effect}'' (CPE) states that 
among all temperized posteriors $p(\btheta|\mathcal{D})^{1/T}$ the best posterior predictive performance on holdout data is achieved at temperature $T < 1$ \citep{wenzel2020good}.
Formally, tempering the posterior corresponds to a $1/T$-scaling of the potential energy function $U(\bm{\theta})$,
\begin{equation}
    p(\bm{\theta} | \mathcal{D})^{1/T} \propto \exp\left(-\frac{1}{T}U(\bm{\theta})\right) \, , \quad \textnormal{where} \,\,\  U(\bm{\theta}) = - \sum_{i=1}^{n}\log p(y_i | \bm{x}_i, \bm{\theta}) - \log p(\bm{\theta}) ,
    \label{eq:posteriorenergy}
\end{equation}
i.e., both the log-likelihood and the log-prior are scaled by ${1}/{T}$. 
For $T = 1$ we have the Bayes posterior, whereas
for $T \to 0$ we obtain a sequence of distributions which have their mass more and more confined around the MAP mode of the distribution \citep{leimkuhler2019partitionedlangevin}. 
We can thus think of the $T \to 0$ limit of posterior inference as MAP estimation.

Note that, besides tempering the posterior as in \eqref{eq:posteriorenergy}, 
one can also temper only the likelihood resp.\ scale only the log-likelihood, 
as is commonly done in VB, see Section~2.3 in \citep{wenzel2020good}.
It is worth pointing out though that both variants are practically equivalent if the prior variance is multiplicative in log-prior pre-factors (such as for Gaussian priors) and if sufficiently many prior variances are grid-searched over as part of the inference pipeline, such that for the best performing posterior-tempered model there is a corresponding likelihood-tempered model with variance scaled by $1/T$ in the grid and vice versa.

There are three main building blocks in Bayesian deep learning that may be at fault for the CPE to emerge:
(i) \emph{model misspecification}: the likelihood model could be misspecified \citep{wenzel2020good, adlam2020cold, aitchison2020statistical, zeno2020cold}, 
(ii) \emph{bad priors}: the priors currently used in deep BNNs may be inadequate \citep{wenzel2020good, fortuin2021bayesian}, or 
(iii) \emph{inaccurate inference}: the inference method might not yield an accurate enough approximation to the true posterior \citep{wenzel2020good, adlam2020cold, fortuin2021bayesian, izmailov2021bayesian}.
Next we review some of the most prominent hypotheses for the emergence of the CPE.

\paragraph{Inaccurate Inference Hypothesis:}

While recent works on the CPE have taken great care to ensure that their SG-MCMC based inference procedure yields as accurate an approximation to the true posterior as possible, 
it remains difficult to definitively assess the approximation accuracy without having access to the (intractable) true posterior.
To rule out obvious problems with the inference mechanism, \cite{wenzel2020good} proposed a set of diagnostics, 
based on comparing ensemble statistics to their theoretically known values.
We have closely monitored these diagnostics in our experiments, however, despite our own extensive efforts to ensure accurate inference, we cannot exclude the possibility that our inference may be inaccurate even though the diagnostics match.

To investigate the inference hypothesis further, together with other foundational questions in Bayesian deep learning, 
\cite{izmailov2021bayesian} recently applied full-batch Hamiltonian Monte Carlo (HMC)\footnote{\cite{izmailov2021bayesian} parallelized the HMC computation over \emph{hundreds} of Tensor Processing Units (TPUs)}, 
which is considered to be the gold-standard in terms of inference accuracy, 
to the models of \cite{wenzel2020good},
showing that with HMC inference and when data augmentation is turned off the CPE disappears. 
While these results can easily be misconstrued as evidence that the approximate inference is at fault in the CPE, 
there are good reasons to think otherwise.

For instance, \cite{izmailov2021bayesian} show, using the code of \cite{wenzel2020good}, that turning off data augmentation alone is sufficient to remove the CPE (cf. Table~7 in \cite{izmailov2021bayesian} Appendix G).
We too can confirm that the CPE does not arise with SG-MCMC based inference applied to \cite{wenzel2020good}'s models when data augmentation is turned off, cf.\ Figure~\ref{fig:svhn_cifar_not_curated_ce} in Section~\ref{sec:experiments}.

From this we can already conclude that either SG-MCMC inference is accurate enough for this specific setting, or inaccurate inference is not necessary for the CPE to emerge\footnote{We can also conclude that data augmentation is sufficient for the CPE to arise (although it is not necessary, as we will see later).}.

A more direct counter argument follows from \citep{adlam2020cold}, 
which have demonstrated that there can be a CPE in Gaussian Processes (GP) regression, where the posterior has a closed form solution, 
provided that the aleatoric uncertainty is overestimated. 
Hence, the CPE can arise in a setting where exact inference is possible.
They also provide experimental evidence that the CPE can arise in classification tasks in the infinite-width neural network Gaussian process (NNGP) limit. 
Thus, while inaccurate inference may be sufficient, it does not appear to be necessary for the CPE to emerge. 
Note that a similar observation was made in \citep{grunwald2012safe, grunwald2017inconsistency}, which demonstrated benefits of tempering, with $T>1$ in their setting, in the context of exact inference.

\paragraph{Curation Hypothesis (\emph{model misspecification}):}

\cite{aitchison2020statistical} devises a theory that attributes the effectiveness of tempering to the fact that standard benchmark datasets such as CIFAR-10 are carefully curated and that we should take this curation into account by tempering BNNs. 
In \cite{aitchison2020statistical}'s model of curation (actual dataset curation may differ), a datapoint $\bm{x}$ is added to the dataset if and only if \emph{all} $S$ labellers independently agree on the label $y_s$ to be assigned to $\bm{x}$, while the datapoint is discarded if at least one pair of labellers $s, s'$ disagrees $y_s \neq y_{s'}$.
The main argument put forward by \cite{aitchison2020statistical} is that if we a priori know that the dataset is curated in the sense described above, we should take this into account in our likelihood model.
The proposed likelihood to be used in the case of curation should then be of the following form
\begin{equation}
    p(y,\bm{x} | \bm{\theta}) \propto p(\{ y_s = y\}_{s=1}^S | \bm{x}, \bm{\theta}) = \prod_{s}p(y_s = y | \bm{x}, \bm{\theta}) \;. 
\end{equation}
Thus, assuming that the labellers are i.i.d., the probability of consensus on label $y$ is 
\begin{equation}
    p(y,\bm{x} | \bm{\theta}) \propto p(y | \bm{x}, \bm{\theta})^S \;,
\end{equation}
which corresponds to a cold posterior where only the log-likelihood is re-scaled while the prior is not (cf.\ discussion at the beginning of this Section). 
\cite{aitchison2020statistical} argues that we should observe $S \approx {1}/{T}$, i.e., the optimal temperature should roughly\footnote{
To obtain an exact correspondence, we would need access to the discarded datapoints in order to be able to marginalize them out, 
cf.\ ``marginalise over unknown latents'' in Section~3 in \citep{aitchison2020statistical} for how to account for the discarded images in a proper Bayesian sense} 
be inversely proportional to the total number of labellers involved in the curation of a datapoint.

Note that \cite{aitchison2020statistical}'s model of curation only includes a data point if \emph{all} labellers agree, but in practice, datasets are often collected with some tolerance of labeller disagreement, e.g.\ in that a datapoint is included if a  certain fraction of labellers agree.
However, a more realistic (weaker) model of curation, that filters out fewer ``hard'' instances from the pool of uncurated datapoints, would only give rise to a weaker, less pronounced CPE: 
if we do not observe the CPE for \cite{aitchison2020statistical}'s simplistic model of curation, which is the most extreme form of curation imaginable, we would not expect to observe it under more realistic models of curation either.
See \secref{sec:dataset_curation_explain} in the Appendix for additional details on how popular datasets like CIFAR-10 or SVHN were collected.

Finally, we would like to point out that 
\cite{adlam2020cold} made a similar, albeit somewhat more general argument regarding curation resp.\ mismatch of aleatoric uncertainty. 
Specifically, they show that the CPE can arise if the model overestimates the aleatoric uncertainty, which is naturally reduced when the dataset is curated.

\paragraph{Data Augmentation Hypothesis (\emph{model misspecification}):} 

Current deep learning practices use a number of techniques, including data augmentation, that technically do not obey the likelihood principle (see Appendix~K in \citep{wenzel2020good} for an in-depth discussion of so-called ``dirty likelihoods'').
The data augmentation hypothesis specifically says that the CPE is largely an artifact of using data augmentation and that turning off data augmentation  is sufficient to remove the CPE.

The hypothesis has recently gained traction with both \cite{izmailov2021bayesian} and \cite{fortuin2021bayesian} pointing out that turning off data augmentation is sufficient to remove the CPE in \cite{wenzel2020good}'s ResNet CIFAR10 setting (cf. Table~7 in \cite{izmailov2021bayesian} Appendix G and Figure~A.11 in \cite{fortuin2021bayesian} Appendix~A).
On the other hand, \cite{wenzel2020good}'s CNN-LSTM IMDB model already had a clean likelihood function and still gave rise to a CPE.
From these observations we can already conclude that data augmentation is sufficient but not necessary for the CPE to arise.

As the performance of deep neural networks is often significantly better when using some form of data augmentation, it is not really an option to just turn it off in BNNs, while properly accounting for it in Bayesian inference does not seem trivial either.
On the one hand, data augmentation affects the data points that enter the likelihood function.
However, while data augmentation may increase the amount of data seen by the model, that increase is certainly not equal to the number of times each data point is augmented (after all, augmented data is not independent from the original data).
On the other hand, considering data augmentation as a form of regularization (constraining the classification functions to be invariant to certain transformations),
one can argue that it should be represented in the prior \citep{wilk2018learning}. 
It remains an interesting open problem how to properly account for data augmentation in a Bayesian sense.

\paragraph{Bad Prior Hypothesis:}

Isotropic Gaussian priors are the de facto standard for modern Bayesian neural network inference \citep{fortuin2021bayesian}. 
However, it is questionable whether such simplistic priors are optimal and whether they accurately reflect our true beliefs about the weight distributions.
The bad prior hypothesis says that the CPE may only be an epi-phenomenon of a misspecified prior.
The underlying argument is as follows:
in classic Bayesian learning the number of parameters remains small and the prior is quickly dominated by the data.
In contrast, in Bayesian deep learning the model dimensionality is typically on the same order if not larger than the dataset size~\citep{kaplan2020scaling}.
For such large models the prior will not be dominated by the data and will continue to exert an influence on the posterior.
Hence the prior is critical.

The hypothesis has already been put forward by \cite{wenzel2020good}, who reported that the CPE becomes stronger with increasing model dimensionality.
It recently got additional empirical support by \cite{fortuin2021bayesian}, who
found that the CPE can be partially alleviated by using heavy-tailed non-Gaussian priors.
More specifically, they find that for fully connected neural networks (FCNNs), 
heavy-tailed priors can both improve predictive performance and alleviate the CPE.
For convolutional neural networks (CNNs), the CPE can also be removed with heavy-tailed priors,
however, the resulting performance gains are less striking.
On the other hand, the performance of CNNs can be improved with correlated priors, although they no longer appear to alleviate the CPE.

Finally, we note that the prior that ultimately matters is the prior over functions that is induced when a prior over parameters is combined with the functional form of a neural network architecture \citep{wilson2020bayesianperspective, izmailov2021bayesian}.
Still, this does not render the prior over parameters irrelevant, 
as innocent-looking priors may inadvertently be highly informative, for instance placing large prior mass on undesirable functions.


\section{Testing the Relative Influence of the Prior}
\label{sec:test_prior_misspec}

A straightforward way to assess the bad prior hypothesis is to monitor the CPE while continuously trading off the relative influence between the prior term and the likelihood term:
if the CPE becomes stronger as the relative influence of the prior increases, this would be an indication that the prior is poor.
The relative weight of the prior versus the likelihood in Bayesian inference is given by a simple factor: the dataset size $n$. 
To see this, recall the posterior energy function in \eqref{eq:posteriorenergy}.
Note how the log-likelihood is a sum over $n$ datapoints, i.e.\ it scales with the dataset size $n$, while the log-prior is independent of $n$. 
This means that the relative influence of the log-prior vanishes at a rate of $1/n$  compared to the influence of the log-likelihood. 
In other words, the prior will exert its strongest influence for relatively small dataset sizes $n$.

In order to test the relative influence of the prior, 
we devise a simple experiment in which we train BNNs on random sub-samples of different sizes, 
recording the optimal temperature for each value of the dataset size $n$. 
For smaller $n$ the relative importance of the prior with respect to the likelihood is larger, while for larger $n$ the prior has smaller influence on the posterior. 
By varying the dataset size~$n$ we can test the following two hypotheses:

\begin{itemize}[leftmargin=8mm]
    \item If a bad prior causes the CPE, we would expect to see a stronger CPE for smaller dataset sizes.
    \item From the theory of curation \citep{aitchison2020statistical} we would expect that random subsamples of the dataset do not cause a change in the optimal temperature $T^*$, i.e.\ we would expect the same $T^*$ regardless of the dataset size $n$. 
\end{itemize}

Note that when performing the sub-sampling experiments, care must be taken to ensure that the SG-MCMC inference has the same total number of parameter gradient updates across all data set sizes.
In particular, fixing the batch size, we have to increase the cycle length (i.e.\ the number of epochs per cycle) for smaller datasets.
This ensures that the number of samples from the posterior and the overall number of gradient updates are the same across dataset sizes.
Note also that we temper the whole posterior (not just the likelihood) to keep the relative influence of the log-likelihood and log-prior terms the same, cf.\ discussion at the beginning of \secref{sec:coldposteriors}.

Finally, we note that a similar sub-sampling experiment to evaluate different priors was suggested in \citep{atanov2018deep}.
Although the proposed experiment is a rather straightforward method to test the quality of a prior, recent works that study the CPE \citep{adlam2020cold, wilson2019bayesian, aitchison2020statistical, zeno2020cold, fortuin2021bayesian, izmailov2021bayesian} did not perform such analysis, despite the ongoing debate on the role of the likelihood and the prior in the CPE.


\section{CPE: A Symptom with Many Causes?}
\label{sec:experiments}

We use the SG-MCMC implementation of~\cite{wenzel2020good} for all our  experiments\footnote{\url{https://github.com/google-research/google-research/tree/master/cold_posterior_bnn}}. 
In particular, we adopt a cyclical step size schedule and, optionally, layerwise preconditioning. 
We explicitly specify when critical features - e.g. data augmentation - are adopted,  otherwise we refer to Appendix~\ref{sec:exp_details} for a detailed description of the experimental setup, including the hyperparameters and estimates of the compute resources we used. 
For each experiment, we consider six temperature parameters $T \in [10^{-3}, 1]$, where a separate Markov chain was used for each temperature. 
We generally report performance in terms of the test cross-entropy, 
whereas the corresponding accuracy and uncertainty measurements can be found in the Appendix.
Shaded areas in the plots below denote standard errors w.r.t.\ the number of random seeds (three in our case).
We also define the \textit{CPE-ratio} ``CPER'', 
\begin{equation}
    \text{CPER} = {\ell_{T^*}}/{\ell_{T=1}} \,\ \in (0,1] \, ,
\end{equation}
as the ratio between the cross-entropy loss at the optimal temperature $T^*$ versus  $T=1$ (Bayes posterior). 
A low CPER indicates that the performance of the tempered posterior is significantly better than the Bayes posterior. 
We perform experiments on SVHN \citep{netzer2011reading} and CIFAR-10 \citep{krizhevsky2009learning}, both of which are curated (see Appendix \ref{sec:dataset_curation_explain}). We use ResNet-20 neural networks \citep{resnet2016} with Gaussian priors $\mathcal{N}(0,1)$, unless stated otherwise. 
Further results, including for CNN-LSTM on IMDB and MLP on MNIST, can be found in the Appendix.

\begin{figure}[h!]
\centering
\begin{subfigure}[b]{0.4\textwidth}
\includegraphics[width=\textwidth]{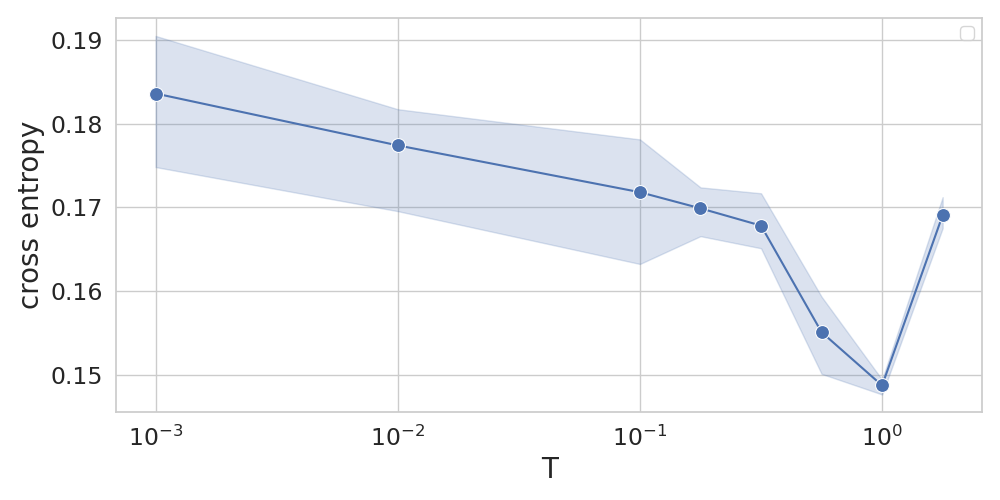}
\caption{SVHN \label{subfig:svhn_cpe_no_data_augm}}
\end{subfigure} \hspace{3mm}
\begin{subfigure}[b]{0.4\textwidth}
\includegraphics[width=\textwidth]{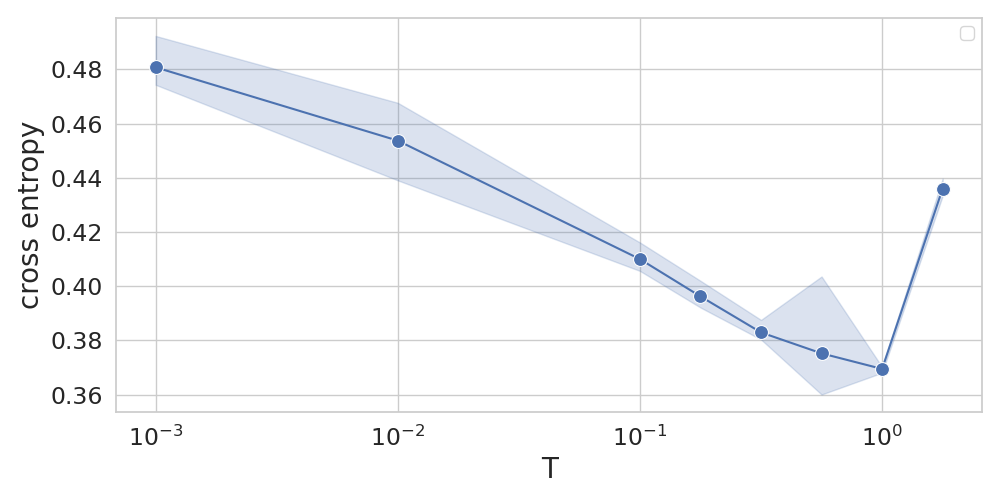}
\caption{CIFAR-10 \label{subfig:svhn_cpe_no_data_augm}}
\end{subfigure}
\caption{Test cross-entropy as a function of temperature $T$ for (a) SVHN and (b) CIFAR-10. 
We can see that $T=1$ is optimal for SVHN and CIFAR-10. }
\label{fig:svhn_cifar_not_curated_ce}
\end{figure}

\paragraph{Starting point: no CPE without data augmentation} 
We run SG-MCMC on SVHN and CIFAR-10 without data augmentation. 
As can be seen in Figure~\ref{fig:svhn_cifar_not_curated_ce}, we observe that SG-MCMC inference on the full dataset $\mathcal{D}$ does not show any sign of a CPE, i.e. $T=1$ is close to optimal. 
From this we can conclude that \emph{either SG-MCMC inference is accurate enough for this specific setting, or inaccurate inference is not necessary for the CPE to emerge}.
We can also conclude that \emph{the curation of SVHN and CIFAR10 does not give rise to a CPE}, which is somewhat surprising from \cite{aitchison2020statistical}'s consensus theory standpoint.
The curation hypothesis is discussed in more detail below.

 \vspace{-2mm}

\subsection{CPE with curation, data augmentation, and random sub-sampling}
\label{sec:CPEwithcurationdataaugmentationandsubsampling}
\begin{tcolorbox}[colback=red!5!white,colframe=red!75!black]
  \textbf{Summary}:  The CPE can arise in case of: synthetic curation, i.e.\ when the the role of the labellers is played by trained neural networks, data augmentation and random sub-sampling. 
\end{tcolorbox}

\paragraph{CPE and synthetic curation}
We now test whether curation can cause the CPE in a simulated environment in which we control the number of labellers and can scale the amount of curation to large levels.
We do so by performing synthetic curation as follows:
\begin{itemize}[leftmargin=8mm]
    \item We first split $\mathcal{D}$ into two non-intersecting sets: a pre-training dataset $\mathcal{D}_{pre}$ and a dataset $\mathcal{D}_{tr}$.
    \item  We train a probabilistic classifier $\hat{\mathcal{S}}$ on $\mathcal{D}_{pre}$, 
    that learns a categorical distribution over the labels (given by the output of the softmax activation of a neural network).
    \item We use $S$ copies drawn from $\hat{\mathcal{S}}$ to independently re-label $\mathcal{D}_{tr}$, effectively simulating the behavior of $S$ i.i.d.\ labellers. 
    The labels are obtained by sampling from the categorical distribution learned by the network. 
    The more uncertain the model is about a prediction for some input, the more disagreement between labellers we expect for that input.
    \item Using the labels induced by $\hat{\mathcal{S}}$, we apply the curation procedure described in \citep{aitchison2020statistical} and summarized earlier in Section \ref{sec:coldposteriors}, to filter $\mathcal{D}_{tr}$ and obtain $\mathcal{D}^{cur}_{tr}$.  Note that the consensus label does not necessarily match with the original label (which can happen for images where the model is confident on the wrong label).
\end{itemize}

As the labeller classifier $\hat{\mathcal{S}}$, we train a ResNet on $\mathcal{D}_{pre}$ with the Adam optimizer. Details of the training procedure can be found in Section~\ref{sec:syntheticcuration} in the Appendix.

\begin{figure}[h!]
\centering
\begin{subfigure}[b]{0.4\textwidth}
\includegraphics[width=\textwidth]{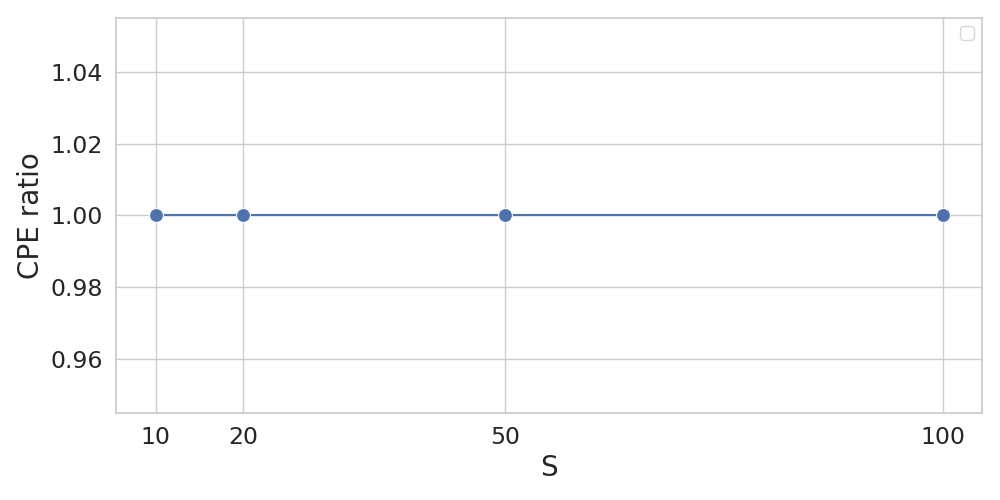}
\caption{Only training set is curated \label{subfig:optimal_temps_only_train}}
\end{subfigure} \hspace{3mm}
\begin{subfigure}[b]{0.4\textwidth}
\includegraphics[width=\textwidth]{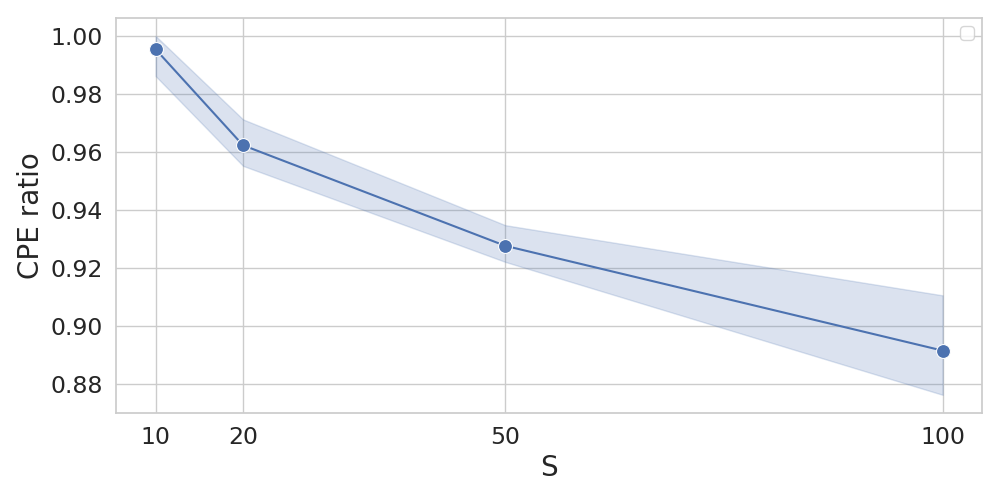}
\caption{Training and test set are curated \label{subfig:optimal_temps_both}}
\end{subfigure}
\caption{CPER as a function of the number of labellers $S$. (a) there is no CPE when only the training set is curated. (b) The CPE arises when the test set is curated, too. 
}.
\label{fig:svhn_labellers_compare}
\end{figure}

We perform SG-MCMC inference on the curated dataset $\mathcal{D}^{cur}_{tr}$ using $S$ labellers, for various values of $S$. 
The cross-entropy is evaluated both on the original test set $\mathcal{D}_{test}$ - which is simply re-labeled according to the trained model $\hat{\mathcal{S}}$ - and the curated one $\mathcal{D}^{cur}_{test}$ using the same number of $S$ labellers as in the training set. 
Surprisingly, \emph{we do not observe any cold posterior effect when only the training set is curated}, as shown in Figure~\ref{subfig:optimal_temps_only_train}. This confirms results of \citep{aitchison2020statistical} (Figure 4D). 
However, as shown in Figure~\ref{subfig:optimal_temps_both}, \emph{curation of both the training and test set causes a CPE}. 

 \vspace{-2mm}

\begin{figure}[h!]
\centering
\begin{subfigure}[b]{0.4\textwidth}
\includegraphics[width=\textwidth]{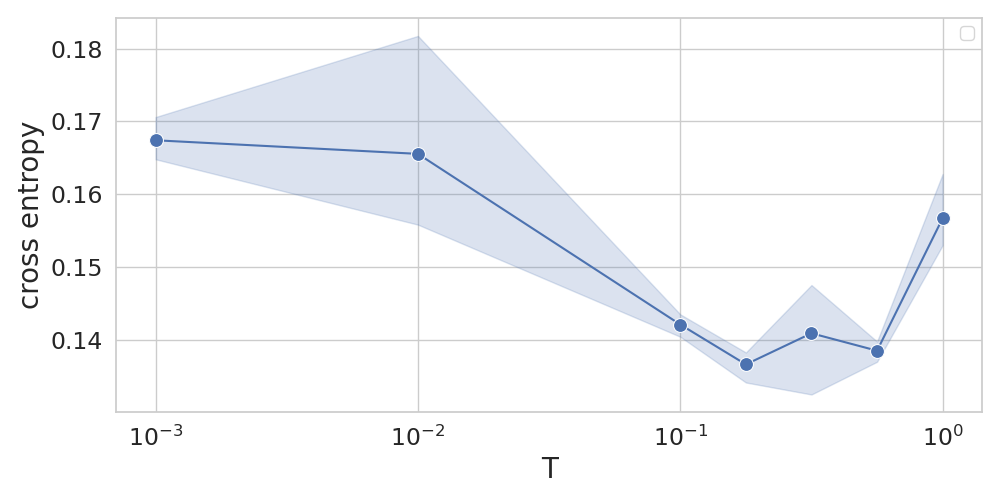}
\caption{SVHN}
\end{subfigure}
\begin{subfigure}[b]{0.4\textwidth}
\includegraphics[width=\textwidth]{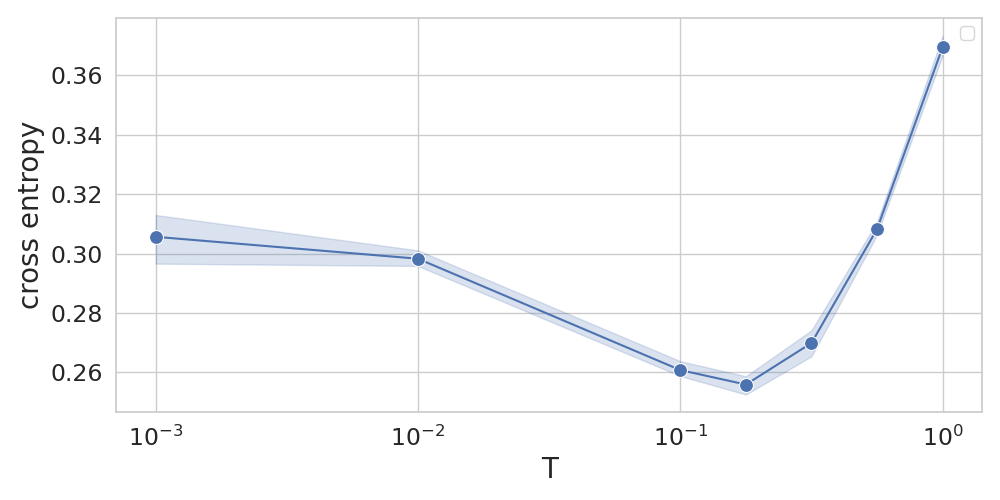}
\caption{CIFAR-10}
\end{subfigure}
\caption{CPE on (a) SVHN and (b) CIFAR-10, both with data augmentation.
}
\label{fig:svhn_cpe_data_augm}
\end{figure}

\paragraph{CPE and data augmentation}
Recent works \citep{izmailov2021bayesian, fortuin2021bayesian}, have identified data augmentation as a cause for the CPE. 
We confirm that data augmentation can cause the CPE on SVHN and CIFAR-10, as shown in Figure~\ref{fig:svhn_cpe_data_augm}. 
In the next paragraph,  
we provide evidence that the CPE can arise even without data augmentation. 
From these observations we can conclude that \emph{data augmentation is sufficient but not necessary for the CPE to arise}.
Note that data augmentation could hint at a problem with the prior, too (when considering data augmentation as a form of regularization).
Details on the kind of data augmentation used for each data set can be found in \secref{sec:dataaugmentationdetails}.

 \vspace{-2mm}

\paragraph{CPE and random sub-sampling}
We now discuss the random sub-sampling experiment to investigate the quality of the prior. 
We run SG-MCMC \emph{without data augmentation} on subsets of SVHN with different sample sizes $n$.
As explained in Section~\ref{sec:test_prior_misspec}, we ensure that the SG-MCMC inference has the same total number of parameter gradient updates across all data set sizes\footnote{We also repeat the sub-sampling experiment in an ablation setting where the number of gradient steps decreases for smaller datasets by reducing the number of epochs per cycle. The results are almost indistinguishable, as can be seen in Section~\ref{sec:furtherexperimentalresults} in the Appendix.}.
The results in Figure~\ref{fig:cpe_vs_dataset_size_same_steps} show that sub-sampling alone can cause CPE. 
Note how the CPE is stronger for smaller~$n$, 
as is evident from the CPE ratio in Figure~\ref{subfig:cpe_ratio_subsample_svhn}. 
As the influence of the prior is larger on smaller datasets, \emph{this is a strong indication that the prior is at fault}\footnote{Note that issues with the prior could also underlie the data augmentation induced CPE observed above: It is difficult to isolate the effect of data augmentation without being able to exclude potential issues with the prior, i.e.\ without knowing for sure what a good prior for BNNs is.}.

\begin{figure}[t!]
\centering
\begin{subfigure}[b]{0.32\textwidth}
\includegraphics[width=\textwidth]{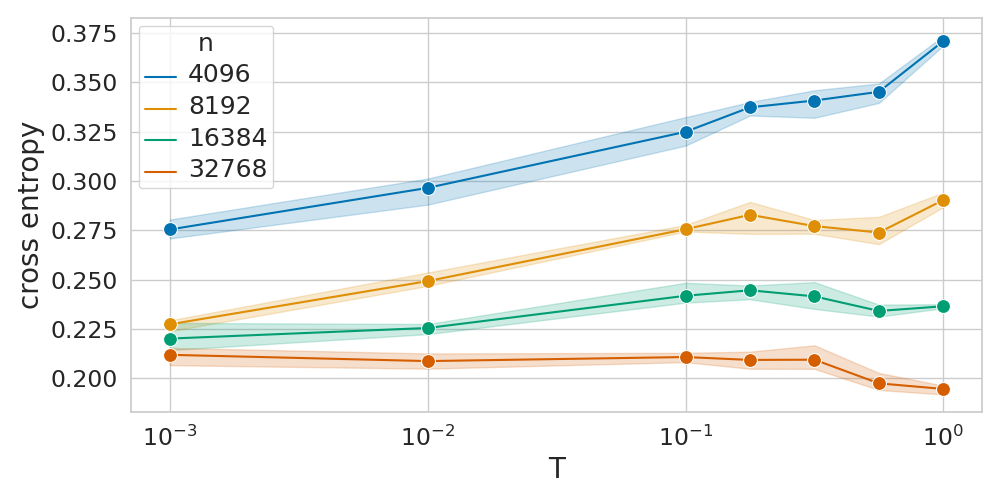}
\caption{ \label{subfig:cpe_subsample_svhn} Cross-entropy}
\end{subfigure}
\begin{subfigure}[b]{0.32\textwidth}
\includegraphics[width=\textwidth]{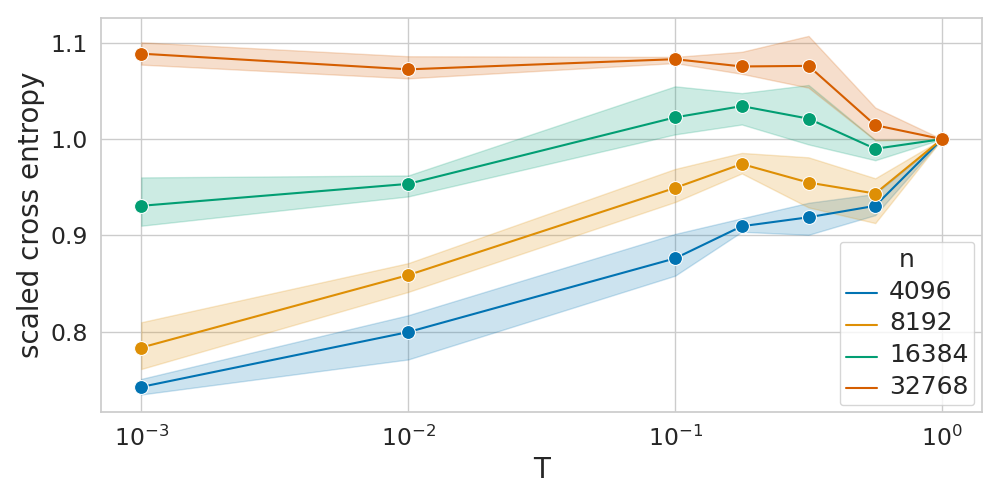}
\caption{ \label{subfig:cpe_subsample_svhn_scaled} Scaled cross-entropy}
\end{subfigure}
\begin{subfigure}[b]{0.32\textwidth}
\includegraphics[width=\textwidth]{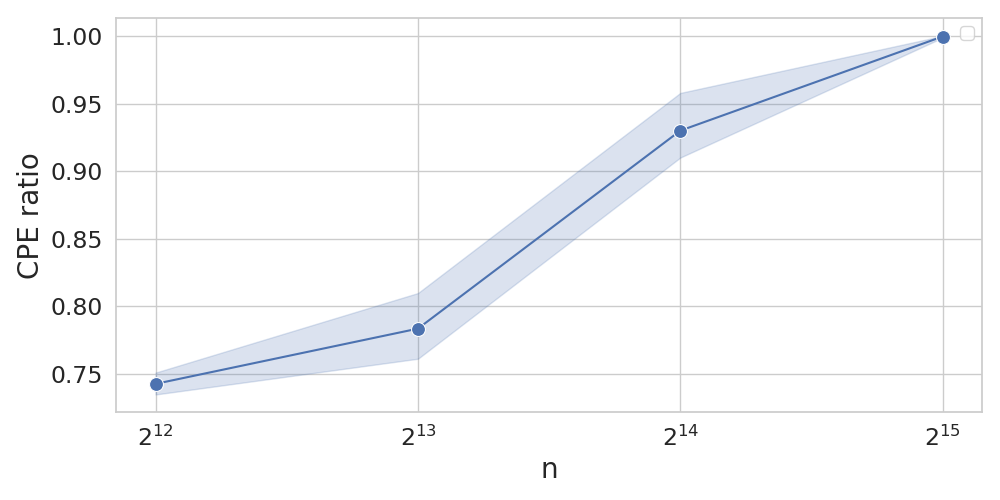}
\caption{ \label{subfig:cpe_ratio_subsample_svhn} CPER}
\end{subfigure}
\caption{SVHN sub-sampling experiment. (a) BNN performance in terms of test cross-entropy for different temperature values. Each line represents a different subsample size, from $n = 2^{12} = 4096$ to $2^{15} = 32768$. (b) same plot, but the test cross-entropy values are scaled by the value at $T=1$. (c) CPER metric as a function of $n$. Note how the CPE is stronger for smaller $n$.}
\label{fig:cpe_vs_dataset_size_same_steps}
\end{figure} \vspace{-2mm}

\subsection{Comparing the relative influence of curation,  data augmentation and sub-sampling}
\label{sec:compare_curation_data_sugm_sub_sampling}

In the previous paragraph, we have seen that sub-sampling alone can cause the CPE to arise.
Here we investigate the effect of sub-sampling on top of curation and data augmentation resp.\ the additional impact of curation and data augmentation on a sub-sampled dataset. 
The precise description of the experimental setup can be found in Section~\ref{sec:relativeinfluence} in the Appendix.
The results are shown in Figure~\ref{fig:svhn_cpe_vs_dataset_size_curated_compare}. 
The plot can be read in two ways: either one looks at a fixed dataset size $n$ and compares the impact of curation (blue) and data augmentation (green) over the original test set (orange), or one looks at the relative change in CPER induced by sub-sampling for a given curve (e.g.\ how much each curve drops when going from $n=16384$ to $n=8192$).

\begin{figure}[h!]
    \centering
    \begin{minipage}{0.36\textwidth}
        \centering
        \includegraphics[width=0.99\textwidth]{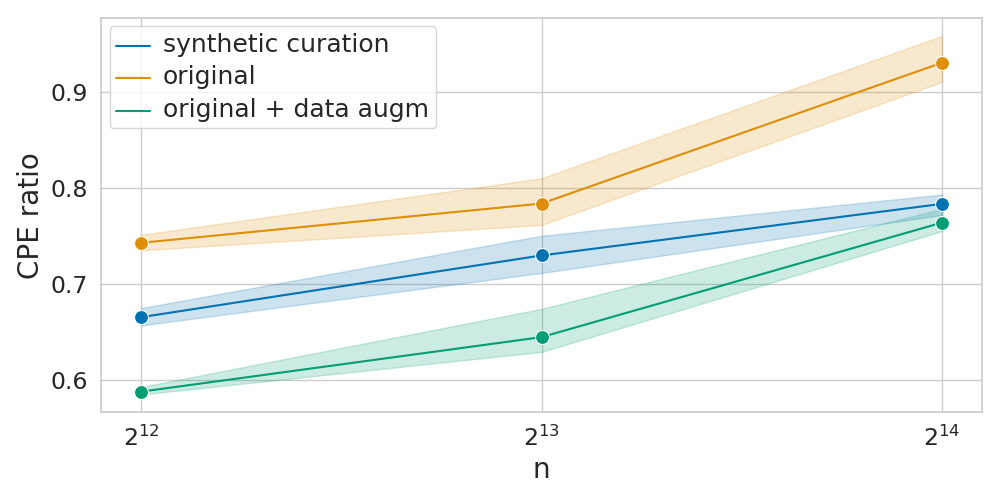} 
        \caption{
        Comparing the CPER for random sub-sampling with curation and data augmentation on datasets of size $4096$, $8192$ and $16384$. 
        }
        \label{fig:svhn_cpe_vs_dataset_size_curated_compare}
    \end{minipage} \hspace{2mm} \hfill
    \begin{minipage}{0.61\textwidth}
        \centering
        \vspace{-2.1mm}
        \includegraphics[width=0.49\textwidth]{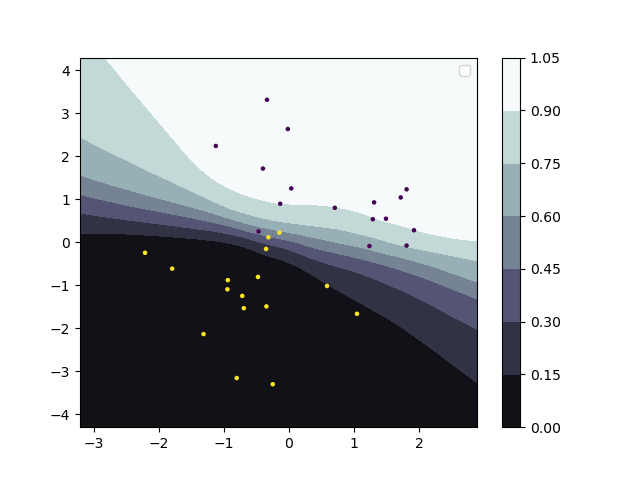} \hspace{-3mm}
        \includegraphics[width=0.49\textwidth]{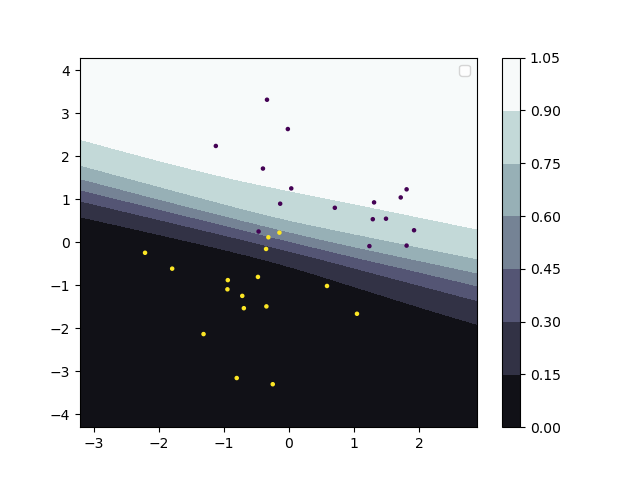}
        \caption{Decision boundary of 1 hidden layer MLP on toy dataset of size $n=32$, (left) $T=1$, (right) $T=10^{-2}$. 
        Color shading according to posterior mean prediction.
        }
        \label{fig:decisionboundary_toyproblem}
    \end{minipage}
\end{figure} \vspace{-2mm}

\subsection{Do standard Gaussian priors give too much weight to complicated hypotheses?}

\begin{tcolorbox}[colback=red!5!white,colframe=red!75!black]
  \textbf{Summary}: 
  The CPE can arise even in a small scale experiment in which a linear separator is optimal. 
  An analysis of the decision boundary suggests that the sharpened posterior obtained through tempered MCMC induces simpler functions than the Bayes posterior at $T=1$.
\end{tcolorbox}
Here we investigate the influence of the prior in a synthetic toy dataset.
To this end, we generate two clusters of datapoints from two 2D Gaussians with variance $\sigma^2 = 1$ centered at $(-1,-1)$ and $(1, 1)$ respectively,
in which the Bayes optimal classifier is given by the straight line $y = -x$.

\begin{figure}[t!]
\centering
\begin{subfigure}[b]{0.32\textwidth}
\includegraphics[width=\textwidth]{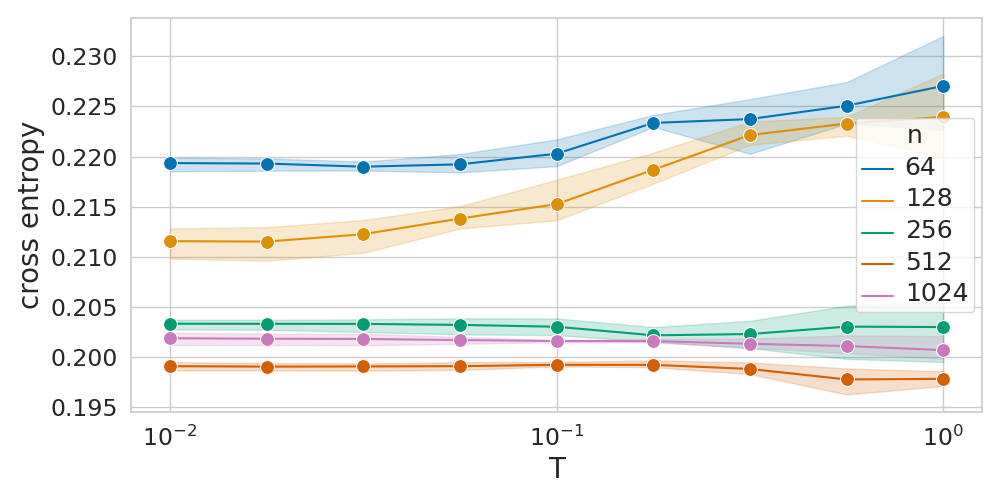}
\caption{Cross-entropy}
\end{subfigure}
\begin{subfigure}[b]{0.32\textwidth}
\includegraphics[width=\textwidth]{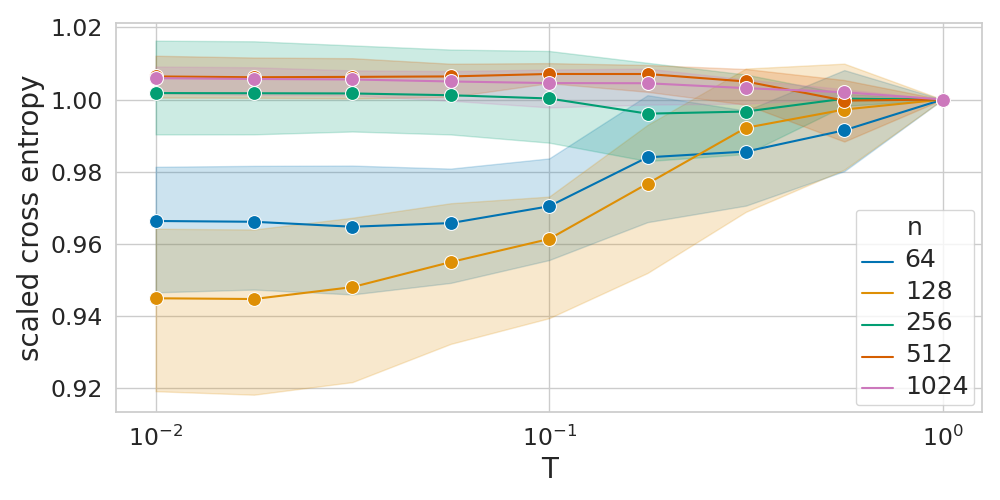}
\caption{Scaled cross-entropy}
\end{subfigure}
\begin{subfigure}[b]{0.32\textwidth}
\includegraphics[width=\textwidth]{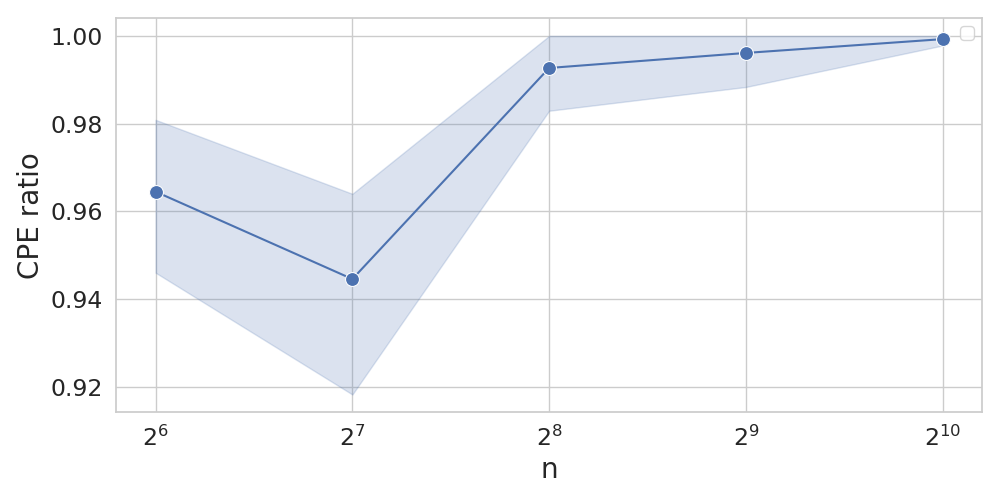}
\caption{CPER}
\end{subfigure}
\caption{Testing the relative influence of the prior on a toy dataset. 
(a) Cross-entropy (CE) as a function of temperature $T$, 
(b) same plot, but the test cross-entropy values are scaled by the value at $T=1$, 
(c) CPE ratio as a function of dataset size~$n$. 
We can see that the CPE becomes stronger for smaller dataset size $n$.
}
\label{fig:cpe_vs_dataset_size_toy}
\end{figure}

To investigate the influence of the prior, we run full-batch MCMC for a 1 hidden layer MLP on subsets of the toy dataset with varying sample sizes $n$ and record the CPE across various temperatures~$T$.
Note that the prior is well-specified in the sense that there is a parameter setting that achieves perfect classification (clearly a 1 hidden layer MLP can learn the optimal separator).
The results, in Figure~\ref{fig:cpe_vs_dataset_size_toy}, show that 
the CPE becomes stronger for smaller dataset size $n$.

An analysis of the decision boundary, shown in Figure~\ref{fig:decisionboundary_toyproblem}, suggests that the sharpened posterior obtained through tempered MCMC induces simpler functions than the Bayes posterior at $T=1$.
We consider it highly relevant future work to investigate if this holds also for real world problems

 \vspace{-2mm}


\section{Discussion \& Conclusion}

The ``\emph{cold posterior effect}'' (CPE) in Bayesian deep learning describes the disturbing observation that the predictive performance of Bayesian neural networks (BNNs) can be significantly improved if the Bayes posterior is artificially sharpened using a temperature parameter $T < 1$.
Since the CPE was identified many researchers have proposed hypotheses to explain the phenomenon. 
However, despite this intensive research effort the effect remains poorly understood.
We have provided novel and nuanced evidence relevant to existing explanations for the CPE.

Our results demonstrate how the CPE can arise in isolation from synthetic curation, data augmentation, and random sub-sampling.
Specifically, we have confirmed that there is no CPE on SVHN and CIFAR-10 if data augmentation is turned off (Figure~\ref{fig:svhn_cifar_not_curated_ce}), which is somewhat surprising from the curation theory standpoint (since both datasets are curated).
On the other hand, we have shown that the CPE can arise when both the training and test set are synthetically curated, i.e.\ when the role of the labellers is played by trained neural networks (Figure~\ref{fig:svhn_labellers_compare}).
Most importantly, we have shown that the CPE can also arise when a dataset that does not show any sign of CPE is randomly sub-sampled (Figure~\ref{fig:cpe_vs_dataset_size_same_steps}), providing a strong indication that the prior is at fault in the CPE (the relative influence of the prior over the likelihood increases with decreasing dataset size).

The CPE may be a symptom with many causes. 
Since many of the recent deep learning advances, such as data augmentation, batch normalization and initialization distributions have been designed specifically for DNNs, it is not surprising that tempering, which gets BNNs closer to DNNs, improves their performance. 
The implication of this is that the Bayesian deep learning community has to find their own ``advances'' specifically tailored to BNNs.

Another conclusion of our work is that priors \emph{do} matter in Bayesian deep learning.
In classic Bayesian learning the number of parameters remains small and the prior is quickly dominated by the data.
In contrast, in Bayesian deep learning the model dimensionality is typically on the same order if not larger than the dataset size.
For such large models the prior will not be dominated by the data and will continue to exert an influence on the posterior.
Moreover, issues with the prior could also underlie some of the other hypotheses, 
e.g.\ to account for data augmentation, which can be considered a form of regularization, we might need to represent it in the prior.
We therefore consider it to be timely to study suitable priors for deep BNNs.

\bibliographystyle{apalike}
\bibliography{bibliography}

\newpage


\appendix


\section{Further Related Work}

\paragraph{``Warm'' Posteriors:}
Motivated by the behavior of Bayesian inference in \emph{misspecified} models~\cite{grunwald2017inconsistency,jansen2013misspecification} extensively studied the so called "generalized" Bayesian inference, i.e, the Bayes posterior in which only the likelihood is tempered. In particular, the "Safe Bayes" framework \citep{grunwald2012safe, grunwald2011safe, grunwald2017inconsistency} was developed to tune the temperature parameter. 
However, these works consider only ``warm posteriors'' $T > 1$ (the inverse temperature is called ``learning rate'' in the relevant literature), as a way to learn under model misspecification. Warm posteriors can arise in a context where the model is misspecified, for instance by assuming homoscedastic noise where the data-generating noise is heteroscedastic. Under this misspecification, the model overfits the datapoints, despite the fact that the prior is well-specified (for instance in their case it is centered around the best performing and non-overfitting solution). 
We hypothesize that in \cite{grunwald2017inconsistency} the prior favours simple models, hence it is beneficial to put more weight onto the prior and use warm posterior. The opposite might happen in BNNs: cold posteriors counteract the effect of a bad prior that tends to prefer overcomplicated solutions. Finally, we mention the work of \cite{bhattacharya2019fractionalposteriors}, in which the authors develop \emph{fractional posteriors} with the goal of decreasing posterior concentration.

\section{Dataset Collection \& Curation}
\label{sec:dataset_curation_explain}

Here we review the way that the two datasets that are mainly used in our experiments, SVHN and CIFAR-10, have been collected and curated. 
\paragraph{SVHN}
The Street View House Numbers dataset \citep{netzer2011reading}, which is divided into a training corpus $\mathcal{D}$ of around $73257$ training images and a test set $\mathcal{D}_{te}$ of around $26k$ images. Although we do not know the exact number of labellers, the dataset has undergone a curation procedure in the sense of  \cite{aitchison2020statistical}. In particular, AMT was adopted\footnote{\url{https://www.mturk.com/}} (quoting from \cite{netzer2011reading}, "The SVHN dataset was obtained from a large number of Street View images using a combination
of automated algorithms and the Amazon Mechanical Turk (AMT) framework").

\paragraph{CIFAR-10}
In CIFAR-10 \citep{krizhevsky2009learning}, labellers followed strict guidelines to ensure high quality labelling of the images. In particular, labellers were instructed that "it's worse to include one that shouldn't be included than to exclude one. False positives are worse
than false negatives", and "If there is more than one object that is roughly equally prominent, reject". The reader is invited to review Appenidx C in \cite{krizhevsky2009learning} .

Unfortunately, in the relevant papers there are no details on the specific curation process that was applied, e.g. the number of labellers per image, or whether \emph{all} labellers have to agree on a label or only a subset of them.

\section{MCMC Inference}
\label{eq:BNNposteriorinference}

In this section we review the basics of (SG)-MCMC inference. The description of the implementation and adaptations for deep learning can be found in Section \ref{sec:inferencemethod} below.

The two main inference paradigms to learn a distribution over model parameters $q(\btheta)\simeq p(\btheta|\D)$ respectively to sample from the posterior $\btheta_{1}, \dots, \btheta_{K} \sim p(\btheta | \D)$ are 
Variational Bayes (VB) \citep{hinton1993mdl, mackay1995ensemblelearning, barber1998ensemblelearning, blundell2015weightuncertainty} and Markov Chain Monte Carlo (MCMC) \citep{neal1995bayesianneuralnetworks, welling2011bayesian, chen2014stochastichmc, ma2015complete, li2016preconditionedsgld}. 
We will focus on MCMC methods as they are simple to implement and can be scaled to large models and datasets 
when used with stochastic minibatch gradients (SG-MCMC) \citep{welling2011bayesian, chen2014stochastichmc}.

Markov Chain Monte Carlo (MCMC) methods allow to sample an ensemble of models $\btheta_{1}, \dots \btheta_{K} \sim p(\btheta | \D)^{1/T}$ from the (tempered) posterior $p(\btheta | \D)^{1/T}$, by performing a guided random walk in parameter space in which artificial noise is injected into the updates $\btheta_{k} \to \btheta_{k+1}$ in such a way that the ensemble distribution converges to the desired posterior $p(\btheta | \mathcal{D})^{1/T}$ (in the limit of small step sizes and long enough run times) \citep{neal1995bayesianneuralnetworks, welling2011bayesian, ma2015complete, li2016preconditionedsgld}.
By injecting artificial noise, the algorithm explores the loss landscape instead of approaching a single point estimate $\widehat\btheta$.

\paragraph{SG-MCMC} Recent advances in stochastic inference through Markov Chain Monte Carlo (MCMC) methods have made the task of sampling from the posterior distribution of deep neural networks more efficient  \citep{welling2011bayesian, zhang2019cyclicalsgmcmc, wenzel2020good}. In particular, the usage of mini-batches gave rise to stochastic gradient MCMC methods (SG-MCMC) \citep{welling2011bayesian}, which is further improved through various techniques such as momentum variables \citep{chen2014stochastichmc}, preconditioning \citep{li2016preconditionedsgld}, and cyclical stepsize \citep{zhang2019cyclicalsgmcmc}. All these methods perform stochastic updates in parameter space that come from the discretization of a stochastic process \citep{ma2015complete}. For the purpose of exposition, here we mention SG-MCMC in its simplest form, given by SGLD (stochastic gradient Langevin dynamics), in which the updates have the form
\begin{equation}
\label{eq:langevin_discr_posterior}
    \bm{\theta}_{t+1} = \bm{\theta}_t + \frac{\epsilon_t}{2} \left( \frac{n}{B}\sum_{i=1}^B \nabla \log p(y | \bm{x}_i, \bm{\theta}) + \nabla \log p(\bm{\theta}) \right) + \bm{\eta}_t ,
\end{equation}
where $B$ indicates the size of a mini-batch, and $\bm{\eta}_t \sim \mathcal{N}(0, \epsilon_t I)$. We define the gradient of the mini-match posterior energy as $\nabla\tilde{U}(\bm{\theta}) := \frac{n}{B}\sum_{i=1}^B \nabla \log p(y | \bm{x}_i, \bm{\theta}) + \nabla \log p(\bm{\theta})$. \cite{welling2011bayesian} show that if $\epsilon_t$ is such that:
\begin{equation}
    \sum_{i=1}^\infty \epsilon_t = \infty  \,\,\ , \quad \sum_{i=1}^\infty \epsilon_t^2 < \infty ,
\end{equation}
then convergence to a local maximum is guaranteed. 
We will use the SG-MCMC implementation of~\citep{wenzel2020good} throughout our experiments\footnote{\url{https://github.com/google-research/google-research/tree/master/cold_posterior_bnn}}, which combines the aforementioned techniques, further discussed in Section \ref{sec:inferencemethod} below.

\section{Experimental Setup}
\label{sec:exp_details}
The experimental details, including the SG-MCMC hyperparameters, are included in Table \ref{tab:exp_details}. Note that for the subsampling experiments on SVHN (Figure~\ref{fig:cpe_vs_dataset_size_same_steps}), the table entries in the last three columns, that have the number of epochs as units (i.e. burn-in period, cycle length, epochs), refer to the \emph{full dataset size}.
When subsampling is applied, the number of epochs are adijusted such that the number of gradient steps is kept fixed. For instance, if half of the dataset is used, the number of epochs, cycle length epochs and burn-in epochs doubles.  
Finally, the experiments are executed on Nvidia DGX-1 GPU nodes equipped with 4 20-core Xeon E5-2698v4 processors, 512 GB of memory and 8 Nvidia V100 GPUs.

\subsection{Neural Network Architectures}
For the SG-MCMC experiments, we use a 20-layer architecture with residual layers \citep{resnet2016} and batch normalization. For the SG-MCMC experiments on the toy dataset, we use a single hidden layer fully connected net with 20 units and ReLU activation function. For MNIST dataset, we use a 3 hidden layers fully connected net with 20 units and ReLU activation function. For the IMDB dataset, we use CNN-LSTM architecture identical to the one used in \cite{wenzel2020good}. The SG-MCMC method that we adopt is the one in \cite{wenzel2020good} and summarized in Sections \ref{sec:inferencemethod} and \ref{eq:BNNposteriorinference}. In particular, no preconditioning is used. The batch size is 128 across all experiments except for the toy dataset experiment, where the batch size equals the dataset size.

\begin{table}
\caption{Architectural features and SG-MCMC hyper-parameters. The double horizontal line splits the experiments of the paper from those in the appendix.}
\label{tab:exp_details}
\centering
\begin{tabular}{p{9em}>{\centering}p{3em}>{\centering}p{3em}>{\centering}p{3em}>{\centering}p{3em}>{\centering}p{3em}>{\centering}p{3em}p{3em}}
\toprule
\multicolumn{8}{c}{Hyper-parameter}                   \\
\cmidrule(r){2-8}
Experiment   &  data augm & batch norm & precond. &  learning rate & burn-in & cycle length & epochs \\
\midrule
SVHN \& CIFAR-10 (Fig. \ref{fig:svhn_cifar_not_curated_ce}) & \xmark & \cmark & \xmark & 0.1 & 150 & 50 & 1500 \\
\midrule
SVHN (Fig.\ref{fig:svhn_labellers_compare}) & \cmark & \cmark & \xmark & 0.1 & 100 & 25 & 500 \\
\midrule
SVHN (Fig. \ref{fig:svhn_cpe_data_augm}) & \cmark & \cmark & \xmark & 0.1 & 150 & 50 & 1500 \\
\midrule
CIFAR-10 (Fig. \ref{fig:svhn_cpe_data_augm}) & \cmark & \cmark & \xmark & 0.1 & 100 & 25 & 500 \\
\midrule
SVHN (Fig. \ref{fig:cpe_vs_dataset_size_same_steps}) & \xmark & \cmark & \xmark & 0.1 & 100 & 25 & 500 \\
\midrule
Toy data (Fig. \ref{fig:cpe_vs_dataset_size_toy}) & \xmark & \xmark & \xmark & 0.1 & 500 & 75 & 2000 \\
\midrule
\midrule
SVHN \& CIFAR-10 (Fig. \ref{fig:svhn_small_full} and \ref{fig:cifar_small_full}) & \xmark & \cmark & \xmark & 0.1 & 100 & 25 & 500 \\
\midrule
MNIST (Fig. \ref{fig:mnist_subsample}) & \xmark & \xmark & \xmark & 0.1 & 150 & 50 & 1500 \\
\midrule
IMDB (Fig. \ref{fig:imdb_subsample}) & \xmark & \xmark & \xmark & 0.1 & 150 & 50 & 1500 \\
\midrule
CIFAR-10H (Fig. \ref{fig:cifar10h_subsample} and \ref{fig:cifar10h_ls_01}) & \cmark & \cmark & \xmark & adapted & 100 & 50 & 1000 \\
\bottomrule
\end{tabular}
\end{table}

\subsection{Inference Method / Training Procedure}
\label{sec:inferencemethod}
In this work, we will mainly use the inference method proposed in \cite{wenzel2020good}, which adapts recent advances in optimization for deep learning and stochastic inference to SG-MCMC. See also \cite{chen2016bridging} for some interesting connections between SG-MCMC and stochastic optimization. 
\paragraph{Momentum Variables}
Adding momentum to SGD is an optimization technique to accelerate gradient based optimization methods that is widely used in deep learning \citep{sutskever2013momentum}. Momentum variables were added to SG-MCMC methods in \cite{chen2014stochastichmc}, giving raise to the stochastic-gradient version of Hamiltonian dynamics (SG-HMC). SGLD can be modified as follows to include them:
\begin{align}
    & \bm{\theta}_{t+1} = \bm{\theta}_t + \epsilon_t \bm{m}_{t+1} \\
    & \bm{m}_{t+1} = (1 - \epsilon_t \alpha)\bm{m}_t - \epsilon_t \nabla \tilde{U}(\bm{\theta}_t) + \sqrt{2}\bm{\eta}_t,
\end{align}
where $\alpha$ is the momentum weight.

\paragraph{Layerwise Preconditioning}
A subset of our experiments were performed both with and without preconditioning, which did not make a big difference.
The reported results are without preconditioning.

\paragraph{Cyclical step size}
Cyclical step size was introduced by \cite{zhang2019cyclicalsgmcmc} to guarantee better exploration, given the fact that posterior exploration is somewhat limited in standard SGLD due to the fact that the learning rate $\epsilon_t$ must be small enough to avoid bias in estimation and MH acceptance/rejection steps. It consists in alternating updates with large learning rate, which allows to overshoot the local minima and therefore having a better \emph{exploration} of the posterior landscape, and updates with very small learning rate, during which samples from the posterior are collected. All the details of the algorithm can be found in \cite{wenzel2020good}, Section 3.

\subsection{Training the synthetic labellers}
\label{sec:syntheticcuration}
For the curation experiment, the role of the labeller is played by a neural network. 

The experiment described in the first paragraph of \secref{sec:CPEwithcurationdataaugmentationandsubsampling} is designed to test whether curation can cause the CPE in a simulated environment in which we control the number of labellers and can scale the amount of curation to large levels.
We do so by performing synthetic curation as follows:
\begin{itemize}[leftmargin=8mm]
    \item We first split $\mathcal{D}$ into two non-intersecting sets: a pre-training dataset $\mathcal{D}_{pre}$ and a dataset $\mathcal{D}_{tr}$.
    \item  We train a probabilistic classifier $\hat{\mathcal{S}}$ on $\mathcal{D}_{pre}$, 
    that learns a categorical distribution over the labels (given by the output of the softmax activation of a neural network).
    \item We use $S$ copies drawn from $\hat{\mathcal{S}}$ to independently re-label $\mathcal{D}_{tr}$, effectively simulating the behavior of $S$ i.i.d.\ labellers. 
    The labels are obtained by sampling from the categorical distribution learned by the network. 
    The more uncertain the model is about a prediction for some input, the more disagreement between labellers we expect for that input.
    \item Using the labels induced by $\hat{\mathcal{S}}$, we apply the curation procedure described in \citep{aitchison2020statistical} and summarized earlier in Section \ref{sec:coldposteriors}, to filter $\mathcal{D}_{tr}$ and obtain $\mathcal{D}^{cur}_{tr}$.  Note that the consensus label does not necessarily match with the original label (which can happen for images where the model is confident on the wrong label).
\end{itemize}

As the labeller classifier $\hat{\mathcal{S}}$, we train an 8-layer ResNet  on $\mathcal{D}_{pre}$ with the Adam optimizer.

We perform SG-MCMC inference on the curated dataset $\mathcal{D}^{cur}_{tr}$ using $S$ labellers, for various values of $S$. 
The cross-entropy is evaluated both on the original test set $\mathcal{D}_{test}$ - which is simply re-labeled according to the trained model $\hat{\mathcal{S}}$ - and the curated one $\mathcal{D}^{cur}_{test}$ using the same number of $S$ labellers as in the training set. 
The network is optimized with the Tensorflow implementation of Adam optimizer \citep{kingma2014adam} using the default parameters.

\subsection{Relative influence of curation, data augmentation and subsampling}
\label{sec:relativeinfluence}

In this Section, we briefly add some additional thoughts and details regarding the setup of the experiment in \secref{sec:compare_curation_data_sugm_sub_sampling}.

If one inspects the core of curation, one may argue that as it consists of \textit{removing} datapoints, the CPE is caused by using smaller datasets rather than the curation procedure itself. 
Therefore, we want to compare the added contribution to the CPE of curation (and data augmentation) with respect to random sub-sampling. 
While adding data augmentation to random sub-sampling is straightforward, in the case of curation it is not practical to find the exact number of labellers $S$ to match a desired dataset size. Instead, we proceed as follows: we choose the number of labellers such that the curated dataset contains around 17k-18k datapoints (the exact number may vary depending on the random seed).\footnote{ As we observe that the classifier $\tilde{S}$ is over confident on many datapoints, we decide to smooth its output distribution, keeping the rank of the probabilities unchanged. We artificially flatten the labelling distribution by raising the probabilities to the power of $\alpha$, where $\alpha < 1$, and then re-normalizing.} Then  we perform random sub-sampling to match the desired dataset size (16384, 8192, 4096 in our case) on the synthetically curated dataset. The only difference with the previous experiments is that we use the original labels, instead of the consensus labels. In this way, if one image $\bm{x}$ belongs to both the synthetically curated dataset and the randomly sampled one, it has the same label $y$. We run SG-MCMC with the same parameters settings as in Section \ref{sec:experiments}, paragraph "CPE and random sub-sampling".

\section{Further Experimental Results}
\label{sec:furtherexperimentalresults}

\subsection{Subsampling on MNIST and IMDB}
\begin{figure}[h!]
\centering
\begin{subfigure}[b]{0.32\textwidth}
\includegraphics[width=\textwidth]{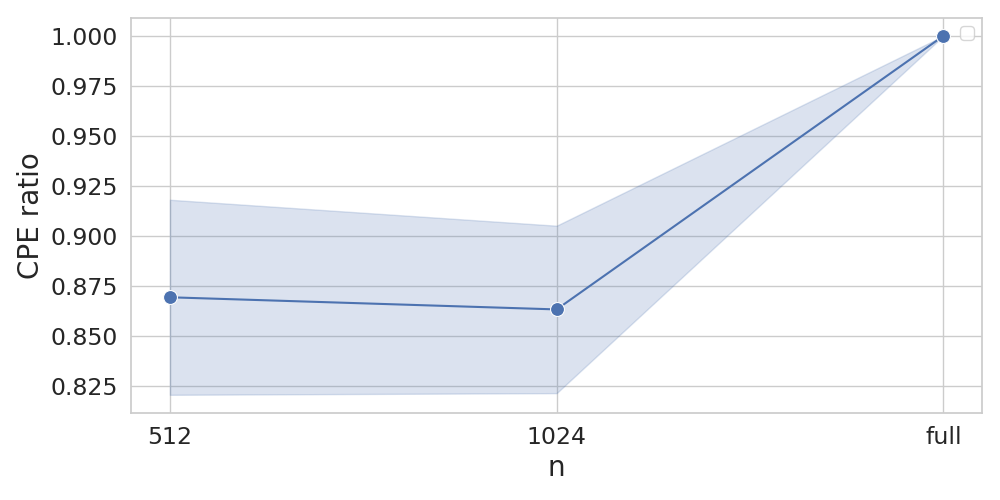}
\end{subfigure}

\begin{subfigure}[b]{0.32\textwidth}
\includegraphics[width=\textwidth]{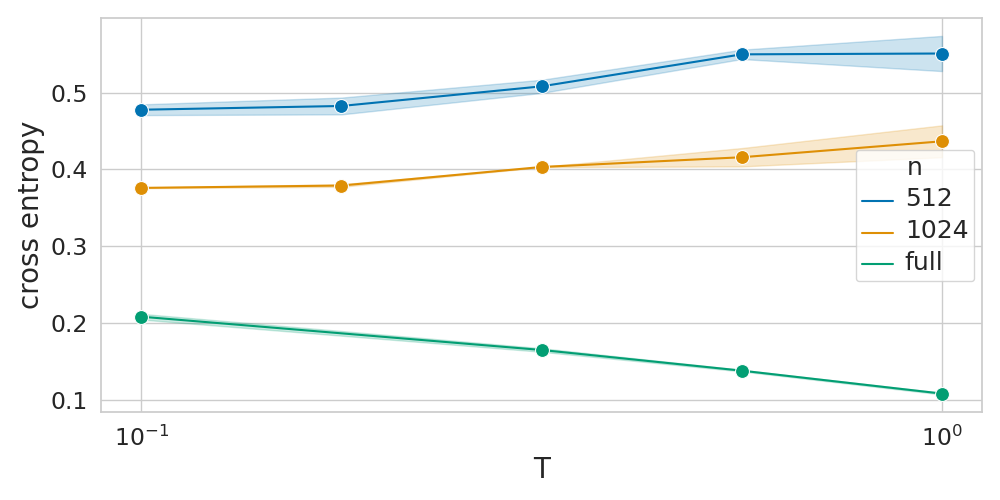}
\end{subfigure}
\begin{subfigure}[b]{0.32\textwidth}
\includegraphics[width=\textwidth]{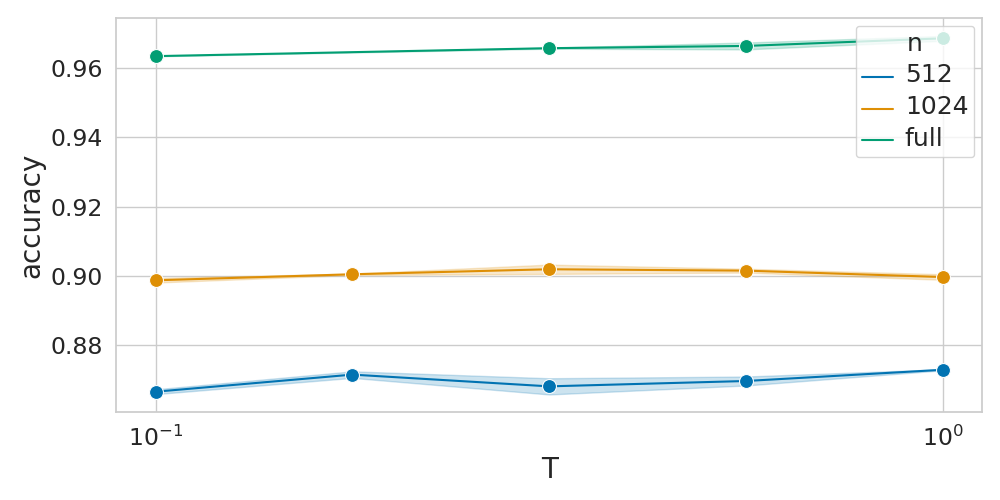}
\end{subfigure}
\begin{subfigure}[b]{0.32\textwidth}
\includegraphics[width=\textwidth]{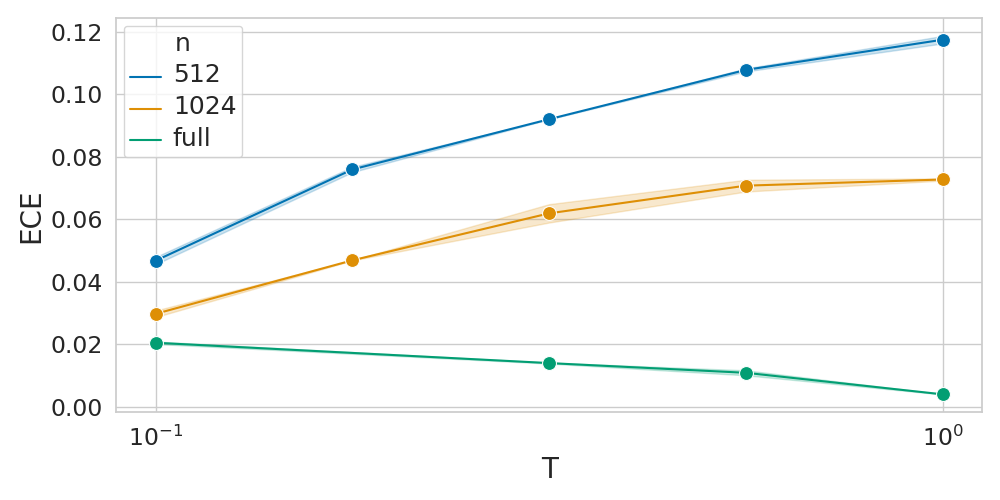}
\end{subfigure}
\caption{MNIST subsampling experiment}
\label{fig:mnist_subsample}
\end{figure}

\begin{figure}[h!]
\centering
\begin{subfigure}[b]{0.32\textwidth}
\includegraphics[width=\textwidth]{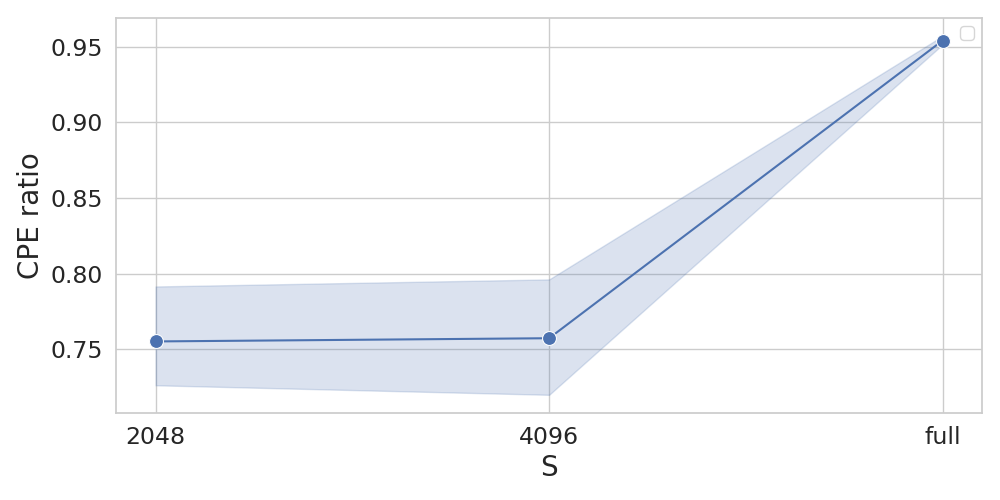}
\end{subfigure}

\begin{subfigure}[b]{0.32\textwidth}
\includegraphics[width=\textwidth]{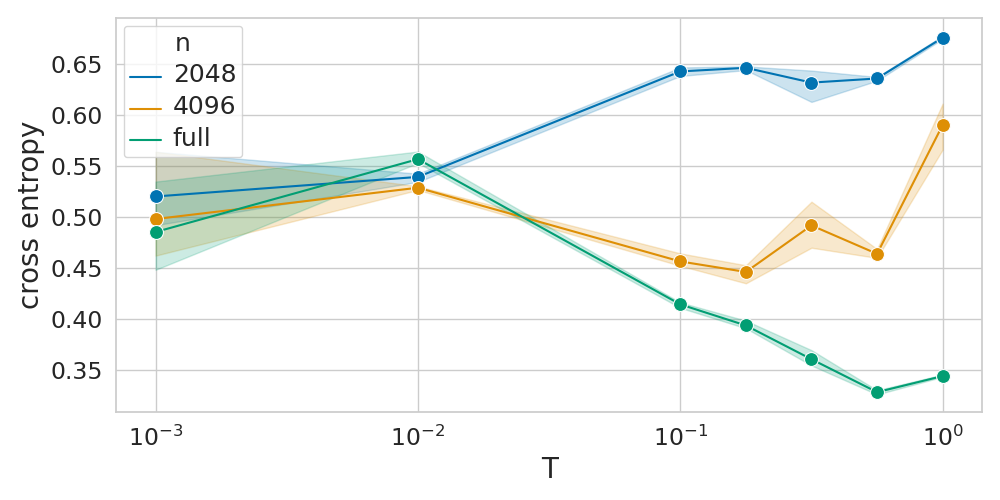}
\end{subfigure}
\begin{subfigure}[b]{0.32\textwidth}
\includegraphics[width=\textwidth]{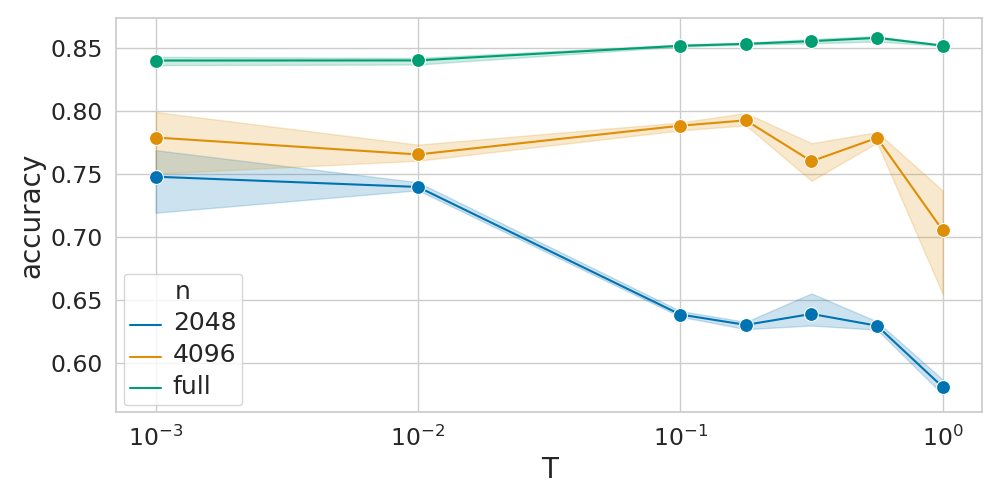}
\end{subfigure}
\begin{subfigure}[b]{0.32\textwidth}
\includegraphics[width=\textwidth]{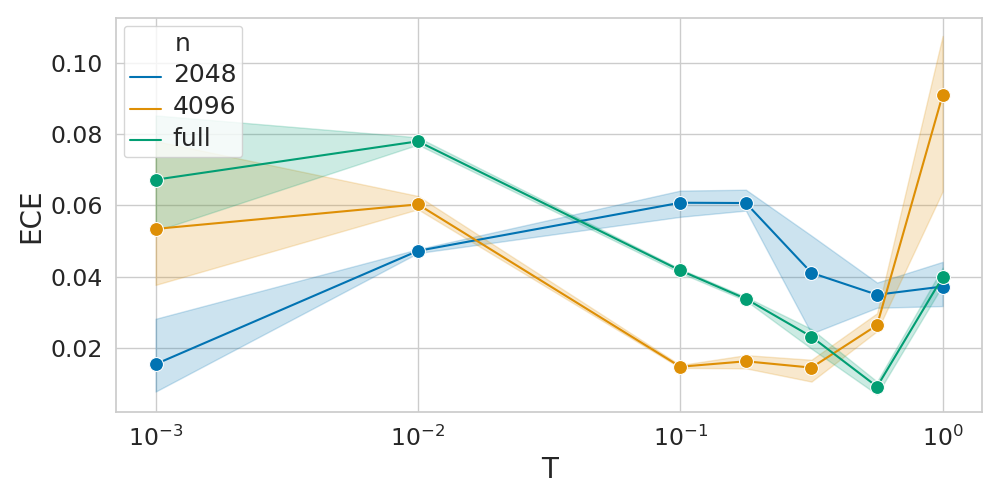}
\end{subfigure}
\caption{IMDB subsampling experiment}
\label{fig:imdb_subsample}
\end{figure}

We repeat the subsampling experiment on MNIST. Results are shown in Figure \ref{fig:mnist_subsample}. Note how for small sample sizes the CPER metric decreases significantly, indicating the presence of the cold posterior effect. The same can be noticed for IMDB (Figure \ref{fig:imdb_subsample}). 

\subsection{CIFAR-10 and SVHN}
We report the values of the expected calibration error (ECE) \cite{guo2017calibration} and accuracy for some selected experiments. For the experiments regarding full SVHN and CIFAR-10 without data augmentation reported in Section \ref{sec:experiments}, Figure  \ref{fig:svhn_cifar_not_curated_ce}, we report these in metrics in Figure \ref{fig:cifar_acc_ece} for CIFAR-10 and 

\begin{figure}[h!]
\centering
\begin{subfigure}[b]{0.32\textwidth}
\includegraphics[width=\textwidth]{cifar10_results/cifar10_cpe.png}
\end{subfigure}
\begin{subfigure}[b]{0.32\textwidth}
\includegraphics[width=\textwidth]{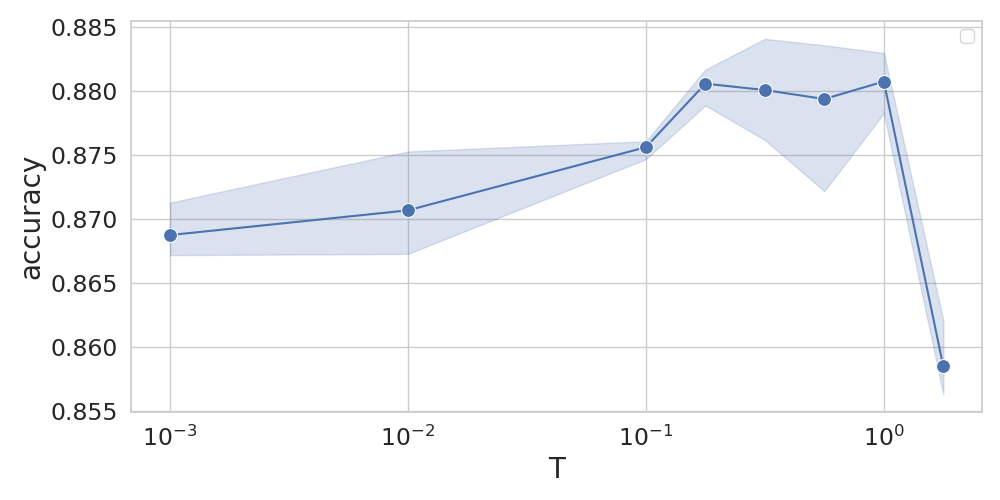}
\end{subfigure}
\begin{subfigure}[b]{0.32\textwidth}
\includegraphics[width=\textwidth]{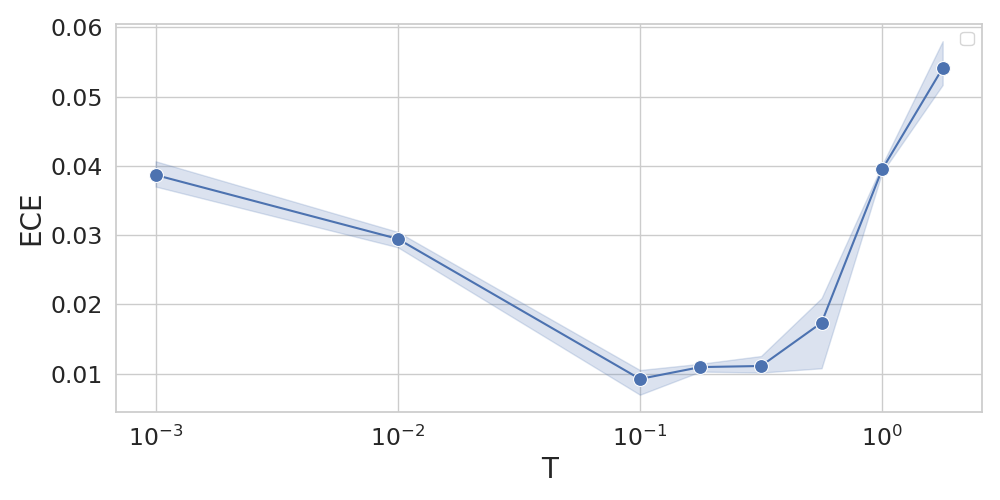}
\end{subfigure}
\caption{CIFAR full dataset experiment (no data augmentation): we additionally report accuracy and ECE.}
\label{fig:cifar_acc_ece}
\end{figure}

\begin{figure}[h!]
\centering
\begin{subfigure}[b]{0.32\textwidth}
\includegraphics[width=\textwidth]{svhn_results/svhn_full_no_data_augm.png}
\end{subfigure}
\begin{subfigure}[b]{0.32\textwidth}
\includegraphics[width=\textwidth]{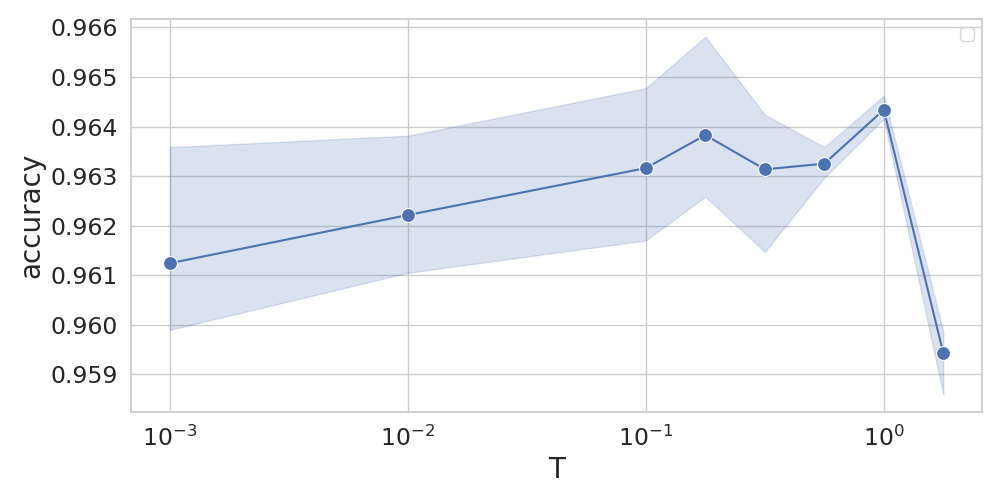}
\end{subfigure}
\begin{subfigure}[b]{0.32\textwidth}
\includegraphics[width=\textwidth]{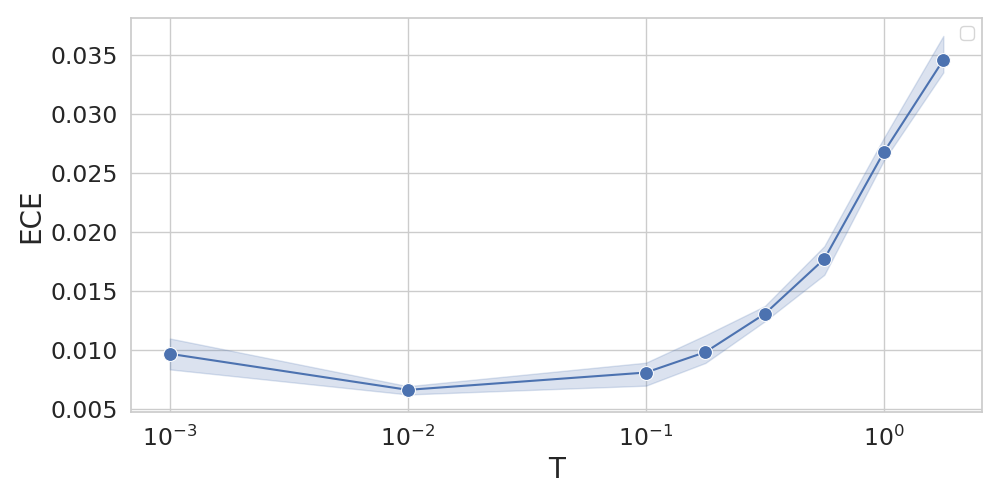}
\end{subfigure}
\caption{SVHN full dataset experiment (no data augmentation): we additionally report accuracy and ECE.}
\label{fig:cifar_acc_ece}
\end{figure}

\begin{figure}[h!]
\centering
\begin{subfigure}[b]{0.32\textwidth}
\includegraphics[width=\textwidth]{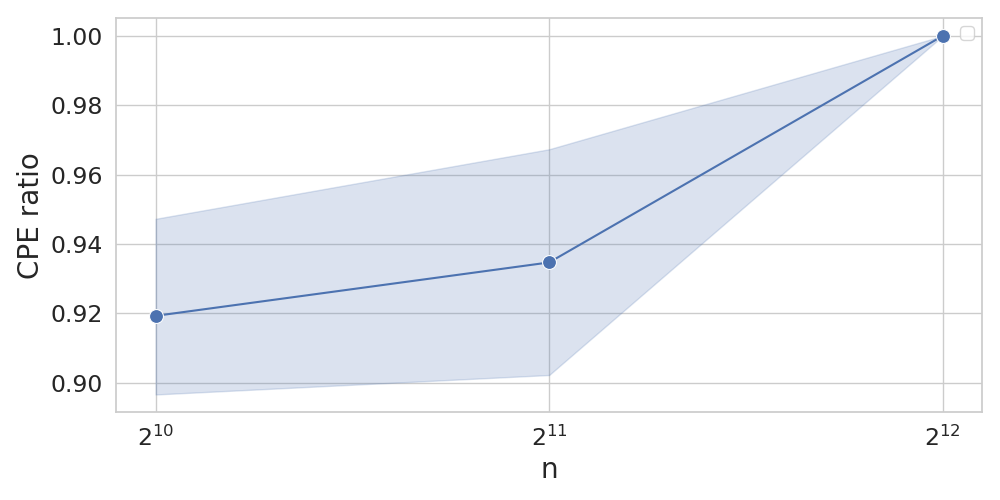}
\end{subfigure}

\begin{subfigure}[b]{0.32\textwidth}
\includegraphics[width=\textwidth]{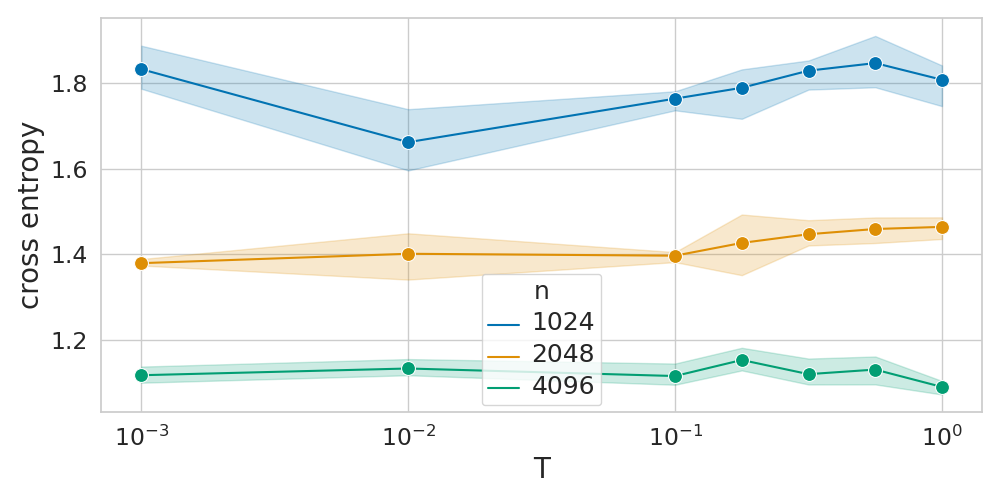}
\end{subfigure}
\begin{subfigure}[b]{0.32\textwidth}
\includegraphics[width=\textwidth]{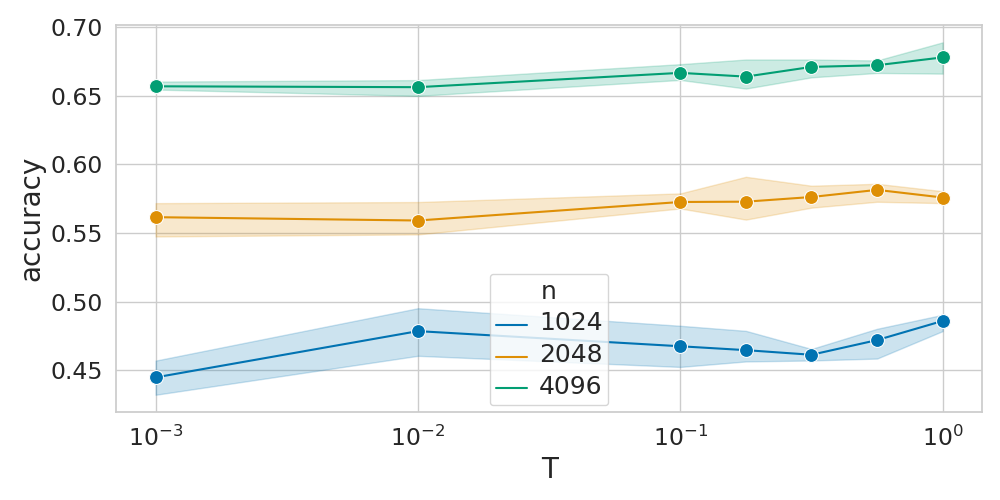}
\end{subfigure}
\begin{subfigure}[b]{0.32\textwidth}
\includegraphics[width=\textwidth]{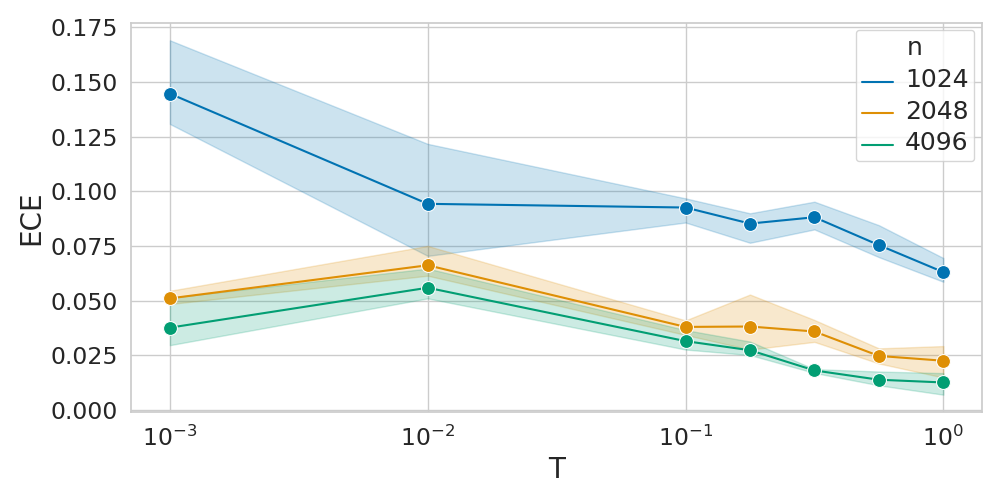}
\end{subfigure}
\caption{CIFAR subsampling experiment: we can see that even for low dataset sizes there is (almost) no CPE on CIFAR-10. In particular, performances are quite insensitive to the temperature $T$.}
\label{fig:cifar_subsample}
\end{figure}

We repeat the subsampling experiment on CIFAR-10. Results are in Figure \ref{fig:cifar_subsample}. For other experiments on full SVHN and CIFAR-10, in which we use less number of epochs per cycle, see Figure \ref{fig:svhn_small_full} for SVHN and Figure \ref{fig:cifar_small_full} for CIFAR-10. 

\begin{figure}[h!]
\centering
\begin{subfigure}[b]{0.32\textwidth}
\includegraphics[width=\textwidth]{svhn_results/cpe_ratio_svhn_no_data_augm_subsample.png}
\end{subfigure}

\begin{subfigure}[b]{0.32\textwidth}
\includegraphics[width=\textwidth]{svhn_results/cpe_svhn_no_data_augm_subsample.png}
\end{subfigure}
\begin{subfigure}[b]{0.32\textwidth}
\includegraphics[width=\textwidth]{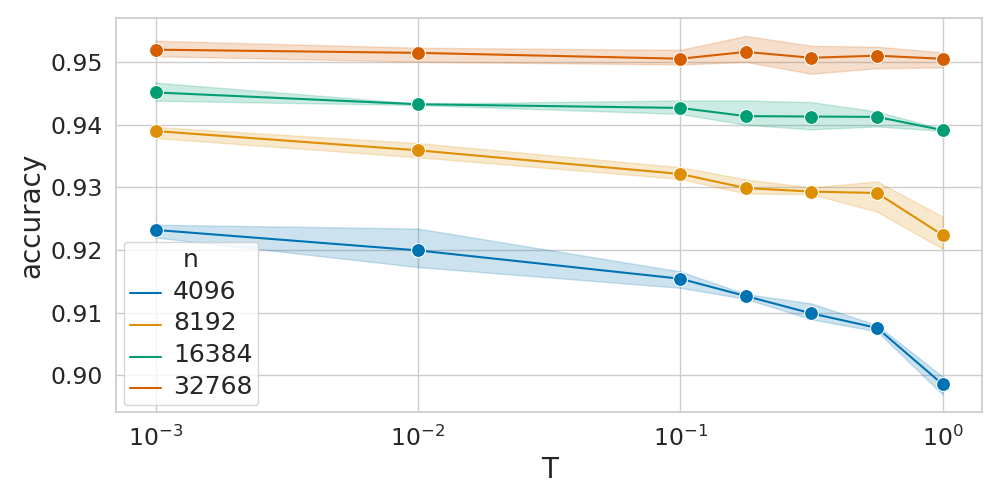}
\end{subfigure}
\begin{subfigure}[b]{0.32\textwidth}
\includegraphics[width=\textwidth]{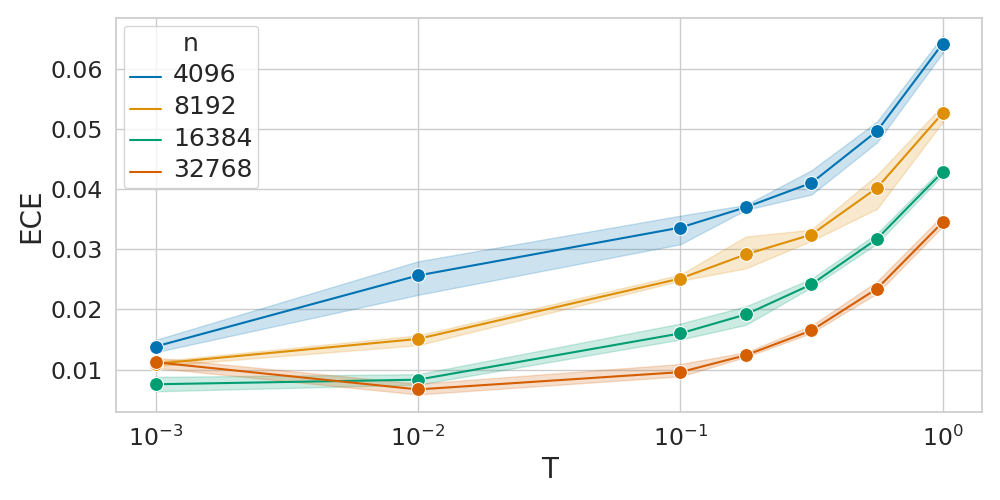}
\end{subfigure}
\caption{SVHN subsampling experiment: the plots for the CPER and cross entropy are the same as Fig. \ref{fig:cpe_vs_dataset_size_same_steps} and added here for completeness.}
\label{fig:svhn_subsample}
\end{figure}

\begin{figure}[h!]
\centering
\begin{subfigure}[b]{0.32\textwidth}
\includegraphics[width=\textwidth]{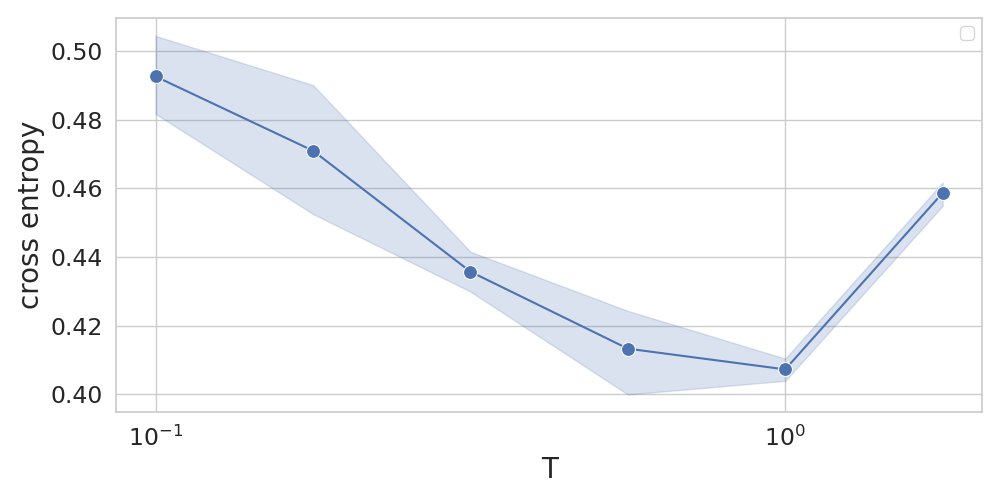}
\end{subfigure}
\begin{subfigure}[b]{0.32\textwidth}
\includegraphics[width=\textwidth]{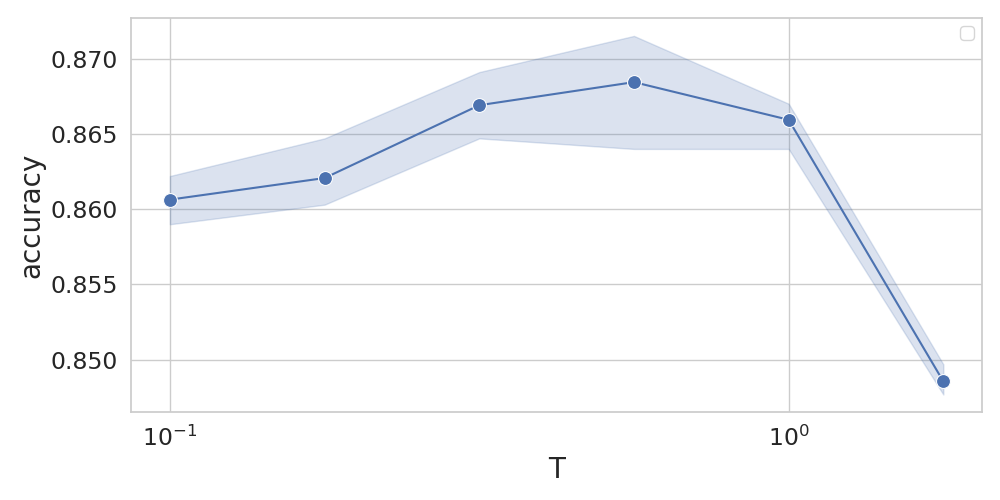}
\end{subfigure}
\begin{subfigure}[b]{0.32\textwidth}
\includegraphics[width=\textwidth]{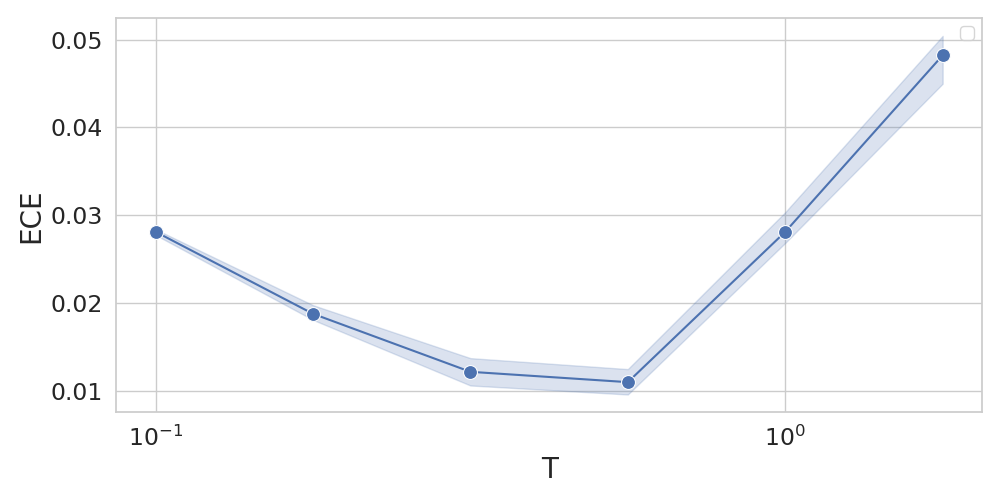}
\end{subfigure}
\caption{CIFAR-10 experiment on full dataset with smaller cycles and less samples (no data augmentation)}
\label{fig:cifar_small_full}
\end{figure}

\begin{figure}[h!]
\centering
\begin{subfigure}[b]{0.32\textwidth}
\includegraphics[width=\textwidth]{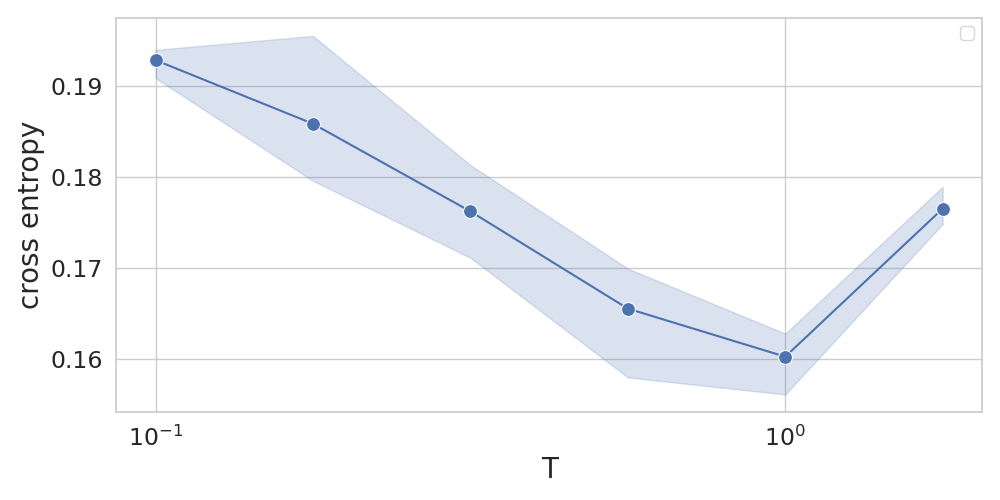}
\end{subfigure}
\begin{subfigure}[b]{0.32\textwidth}
\includegraphics[width=\textwidth]{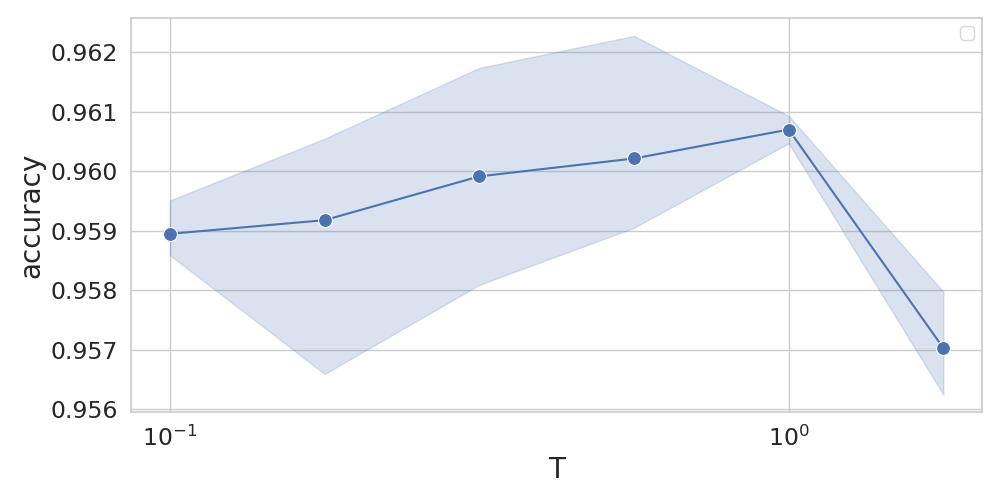}
\end{subfigure}
\begin{subfigure}[b]{0.32\textwidth}
\includegraphics[width=\textwidth]{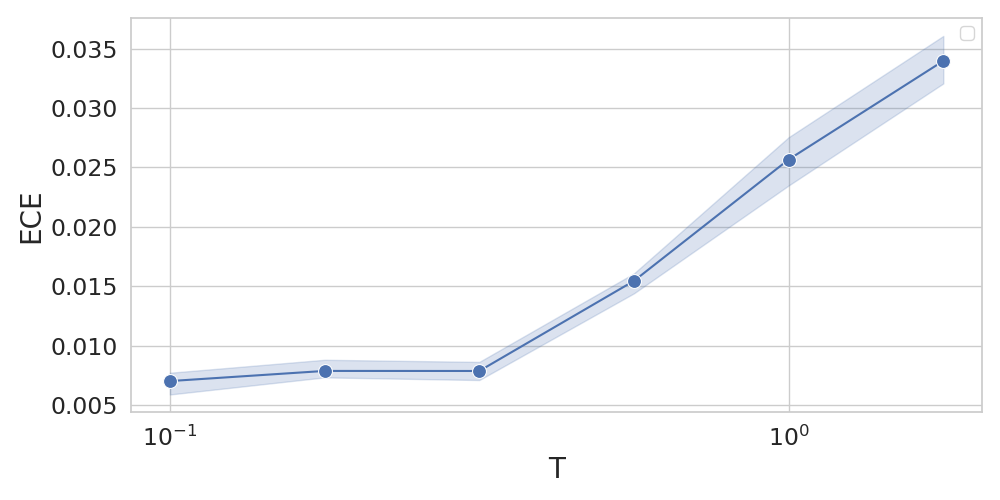}
\end{subfigure}
\caption{SVHN experiment on full dataset with smaller cycles and less samples (no data augmentation)}
\label{fig:svhn_small_full}
\end{figure}

Finally, we additionally report accuracy and ECE for the SVHN subsampling experiments without both data augmentation and curation described in Section \ref{sec:experiments}. See Figure \ref{fig:svhn_subsample}.

\subsection{SVHN curated}
\begin{figure}[h!]
\centering
\begin{subfigure}[b]{0.4\textwidth}
\includegraphics[width=\textwidth]{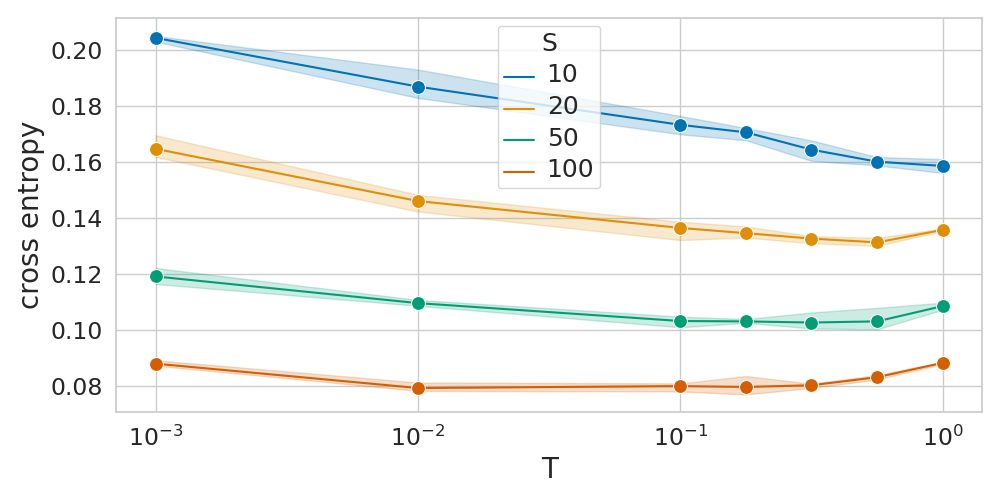}
\end{subfigure}
\begin{subfigure}[b]{0.4\textwidth}
\includegraphics[width=\textwidth]{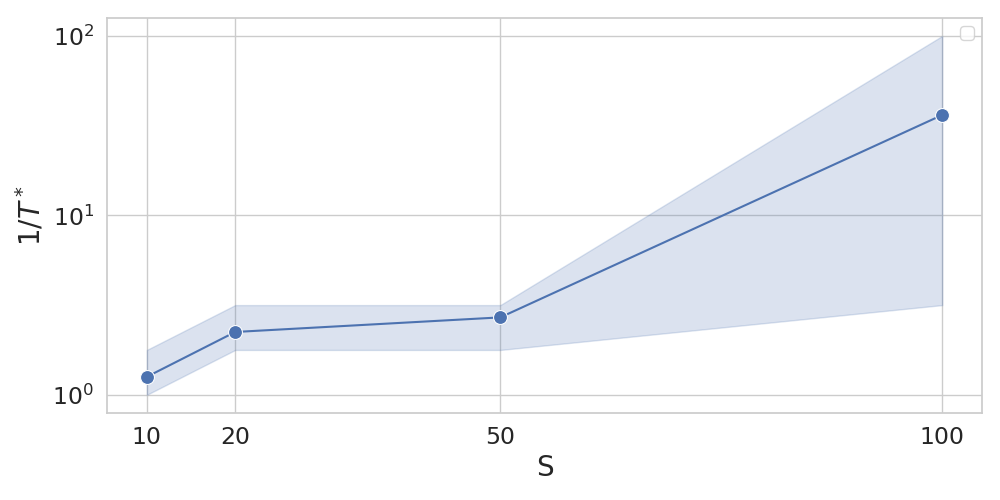}
\end{subfigure}
\caption{Plots underlying experiments when both train and test set are curated}
\label{fig:underlying_plots}
\end{figure}

\begin{figure}[h!]
\centering
\begin{subfigure}[b]{0.4\textwidth}
\includegraphics[width=\textwidth]{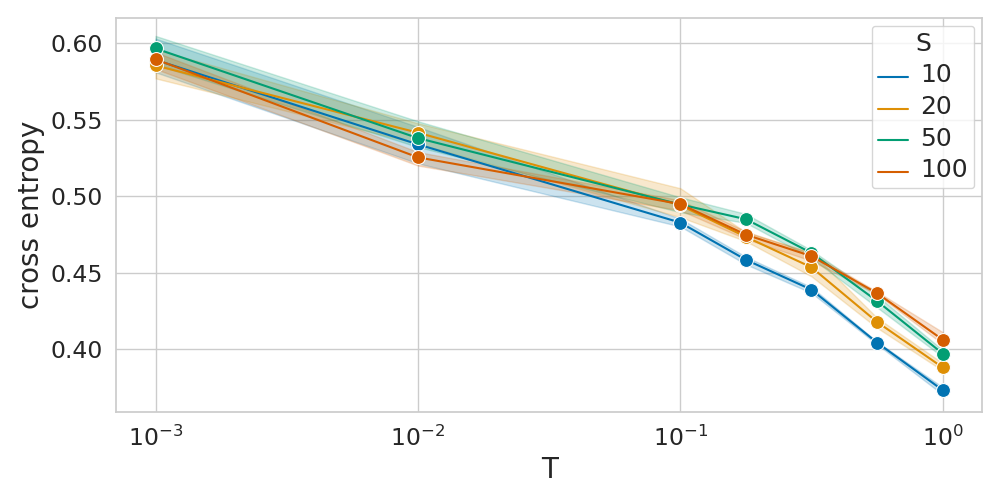}
\end{subfigure}
\begin{subfigure}[b]{0.4\textwidth}
\includegraphics[width=\textwidth]{svhn_results/svhn_only_train_curated_cpe_ratio.png}
\end{subfigure}
\caption{Plots underlying experiments when only the training set is curated: they all perform quite similarly.}
\label{fig:underlying_plots_only_training_set}
\end{figure}

\begin{figure}[h!]
\centering
\begin{subfigure}[b]{0.4\textwidth}
\includegraphics[width=\textwidth]{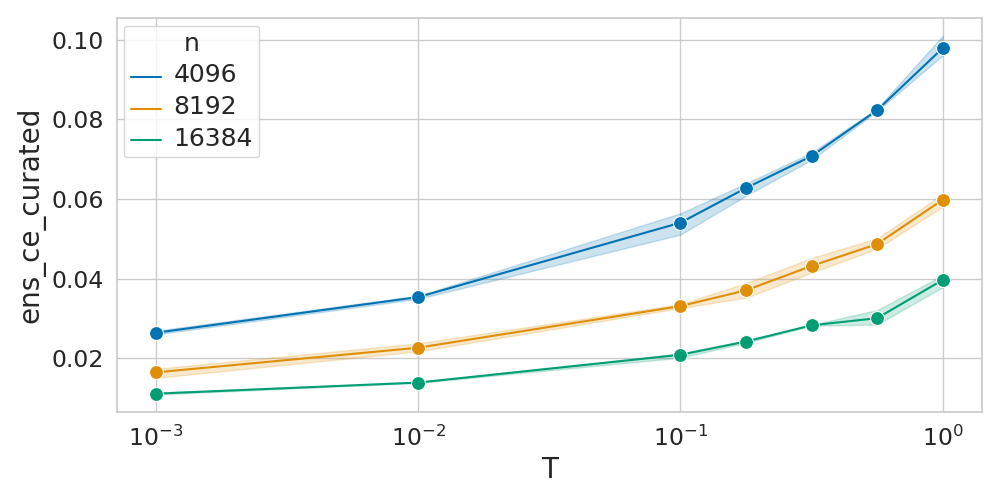}
\end{subfigure}
\begin{subfigure}[b]{0.4\textwidth}
\includegraphics[width=\textwidth]{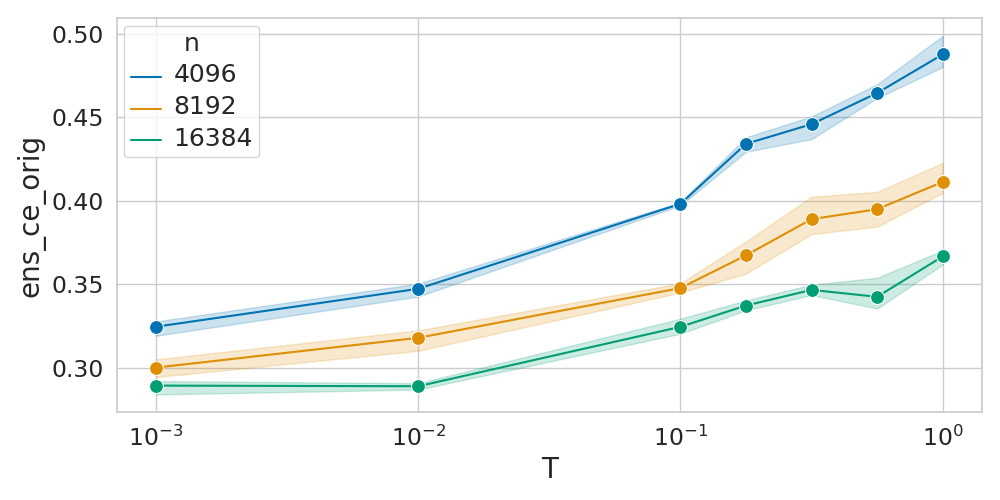}
\end{subfigure}
\caption{Curation + subsample: underlying plots. Left: curated test set. Right: original test set}
\label{fig:curation_subsample_underlying}
\end{figure}

In Figure \ref{fig:underlying_plots} and \ref{fig:underlying_plots_only_training_set}, we show the plots underlying the curation experiment on SVHN, summarized in Figure \ref{fig:svhn_labellers_compare}. In Figure \ref{fig:curation_subsample_underlying} we show the underlying plots of the curation + subsampling experiment of Figure \ref{fig:svhn_cpe_vs_dataset_size_curated_compare}.

\subsection{Data Augmentation}
\label{sec:dataaugmentationdetails}
When data augmentation is used, we perform a random sequence of transformations to every batch during training that causes small class-preserving changes in the images. These transformation, applied at every batch and every epoch, are as follows.

On CIFAR-10 (same as \cite{wenzel2020good}):
\begin{itemize}[leftmargin=8mm]
    \item random left/right flip of the image .
    \item zero padding that expands the size of the input image by four pixels horizontally and vertically, and then random cropping to the original size.
\end{itemize}

On SVHN:
\begin{itemize}[leftmargin=8mm]
    \item adjust the contrast by a random contrast factor between 0.45 and 0.55.
    \item adjust the brightness by adding a factor randomly chosen between -0.15 and 0.15 to each channel.
    \item zero padding that expands the size of the input image by four pixels horizontally and vertically, and then random cropping to the original size.
\end{itemize}

\begin{figure}[h!]
\centering
\begin{subfigure}[b]{0.32\textwidth}
\includegraphics[width=\textwidth]{svhn_results/svhn_full_data_augm.png}
\end{subfigure}
\begin{subfigure}[b]{0.32\textwidth}
\includegraphics[width=\textwidth]{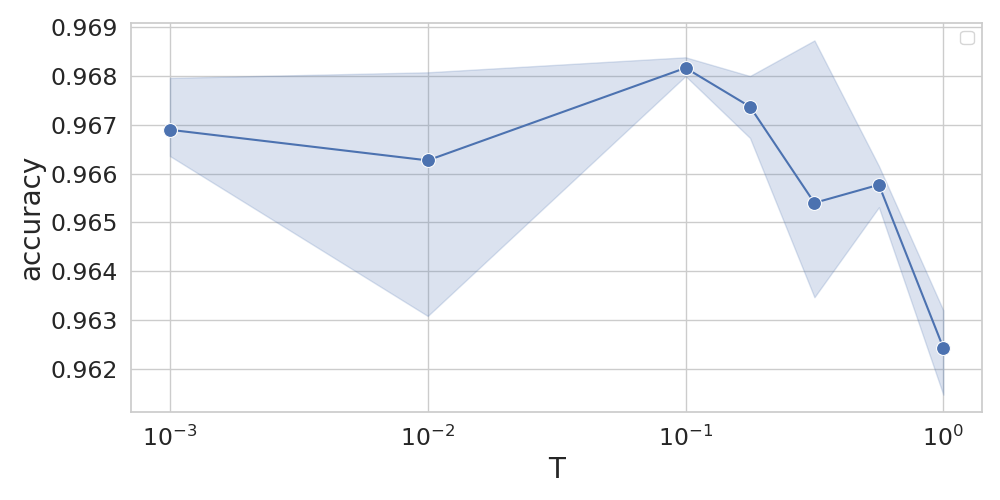}
\end{subfigure}
\begin{subfigure}[b]{0.32\textwidth}
\includegraphics[width=\textwidth]{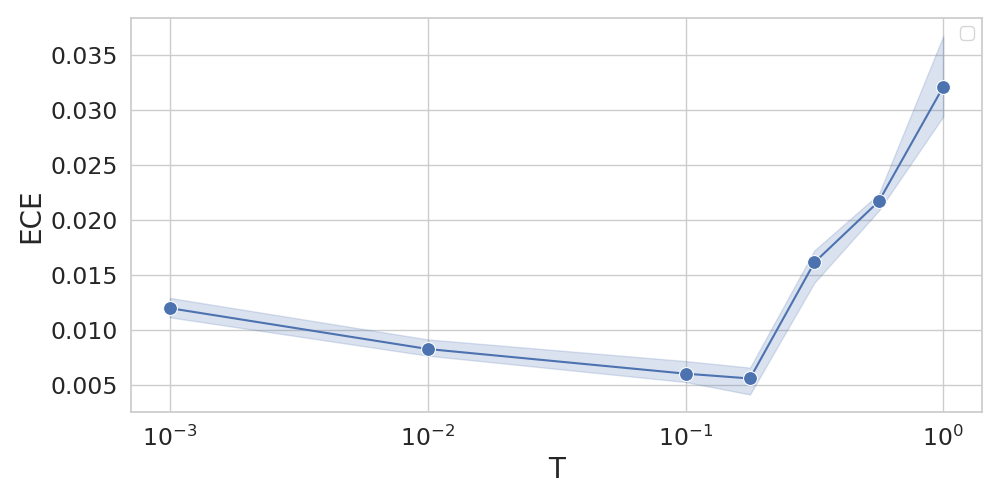}
\end{subfigure}
\caption{CPE on SVHN with data augmentation.
}
\label{fig:svhn_cpe_data_augm_acc_ece}
\end{figure}

\begin{figure}[h!]
\centering
\begin{subfigure}[b]{0.32\textwidth}
\includegraphics[width=\textwidth]{cifar10_results/cifar10_cpe_small_data_augm.png}
\end{subfigure}
\begin{subfigure}[b]{0.32\textwidth}
\includegraphics[width=\textwidth]{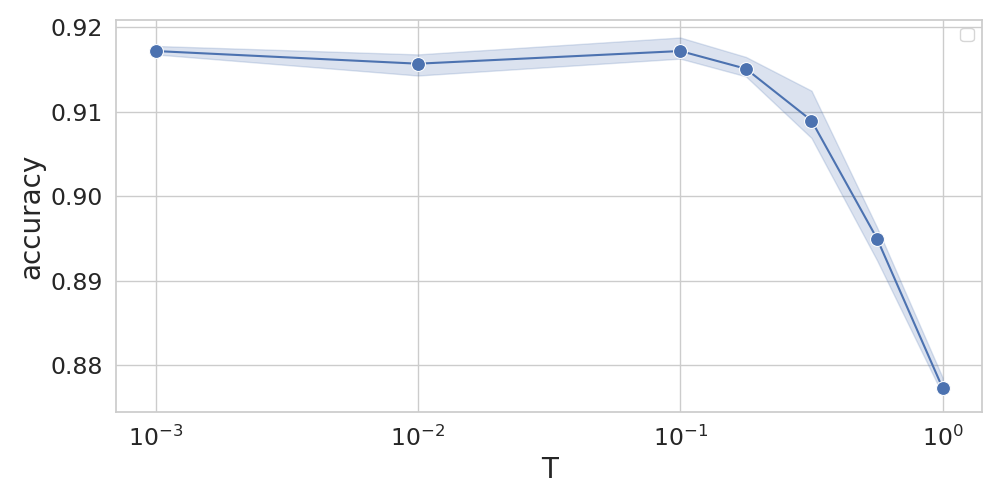}
\end{subfigure}
\begin{subfigure}[b]{0.32\textwidth}
\includegraphics[width=\textwidth]{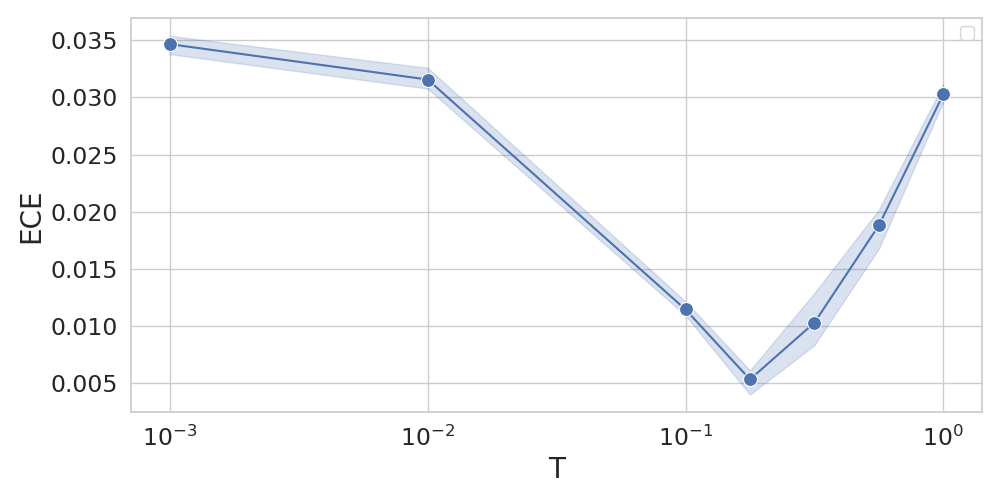}
\end{subfigure}
\caption{CPE on CIFAR-10 with data augmentation.
}
\label{fig:cifar10_cpe_data_augm_acc_ece}
\end{figure}

In Figure \ref{fig:svhn_cpe_data_augm_acc_ece} and \ref{fig:cifar10_cpe_data_augm_acc_ece}, we show the accuracy and ECE for the SVHN experiments with data augmentation of Section \ref{sec:experiments}, Figure \ref{fig:svhn_cpe_data_augm}. 

\begin{figure}[h!]
\centering
\begin{subfigure}[b]{0.32\textwidth}
\includegraphics[width=\textwidth]{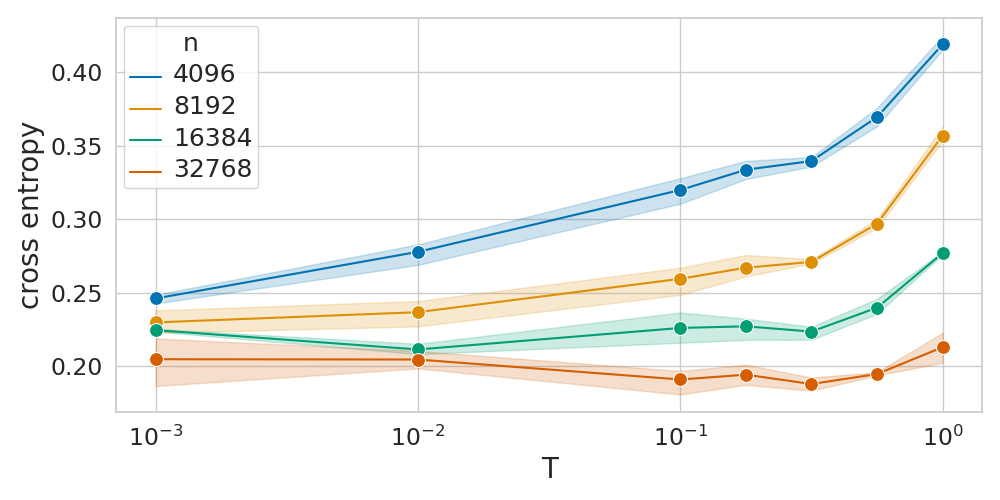}
\end{subfigure}
\begin{subfigure}[b]{0.32\textwidth}
\includegraphics[width=\textwidth]{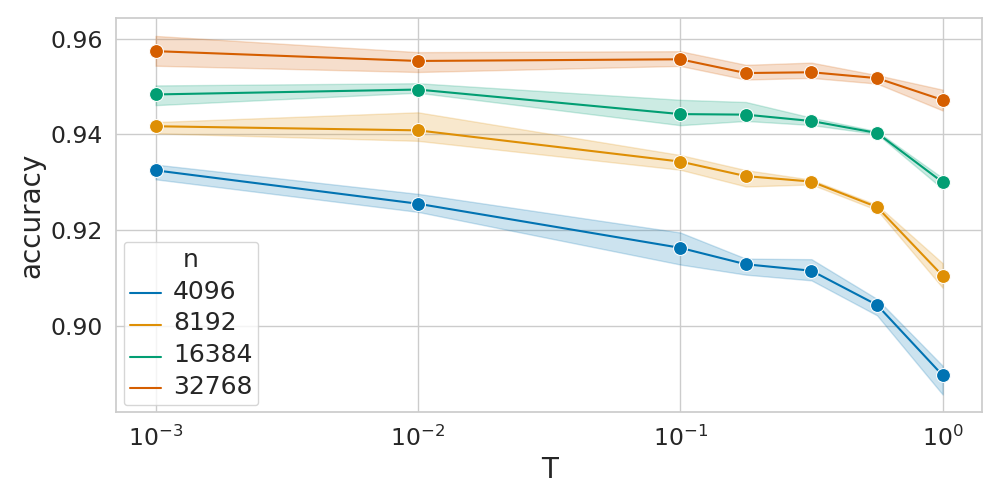}
\end{subfigure}
\begin{subfigure}[b]{0.32\textwidth}
\includegraphics[width=\textwidth]{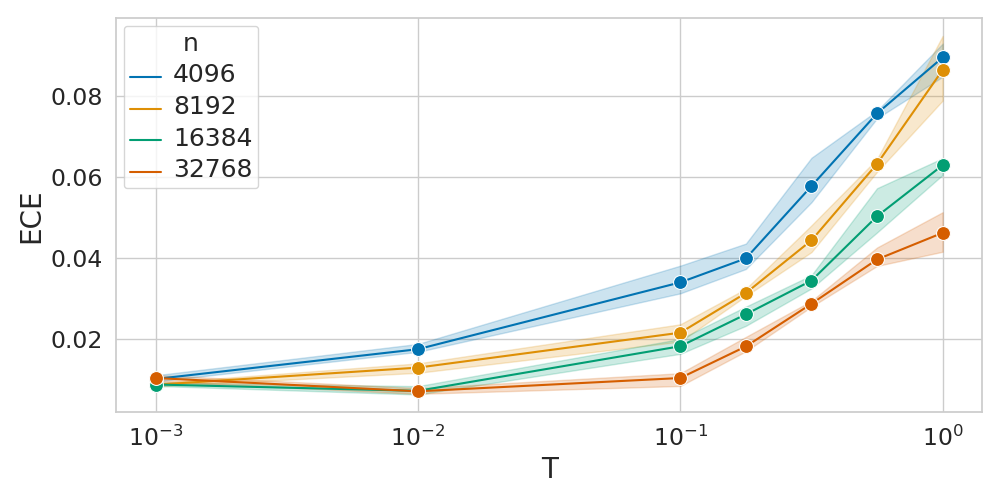}
\end{subfigure}
\caption{Data augmentation and sub-sampling. We additionally show the results for accuracy and ECE. Note a strong cold posterior effect}
\label{fig:plots_underlying_subsample_data_augm}
\end{figure}

In Figure \ref{fig:plots_underlying_subsample_data_augm}, we show the plots underlying Figure \ref{fig:svhn_cpe_vs_dataset_size_curated_compare} for the data augmentation part.

\subsection{Subsampling with decreasing number of gradient steps}
\begin{figure}[h!]
\centering
\begin{subfigure}[b]{0.32\textwidth}
\includegraphics[width=\textwidth]{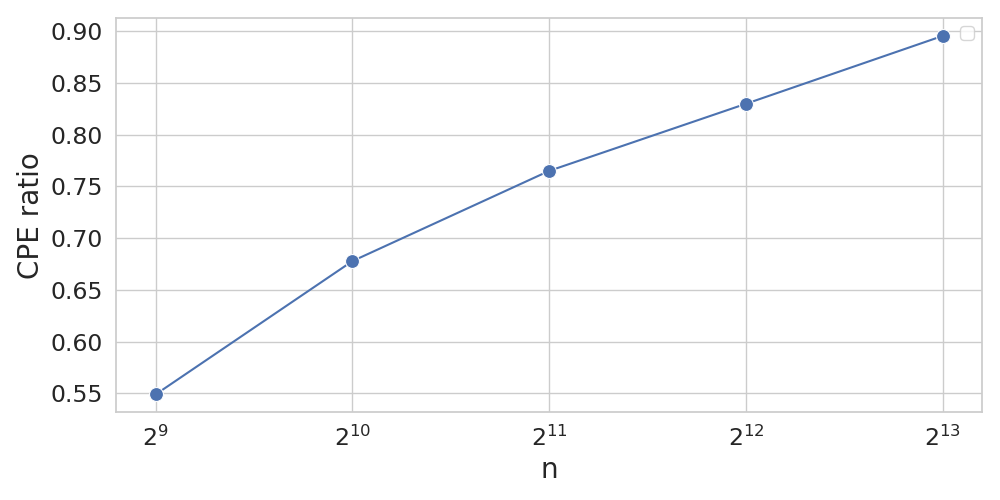}
\end{subfigure}

\begin{subfigure}[b]{0.32\textwidth}
\includegraphics[width=\textwidth]{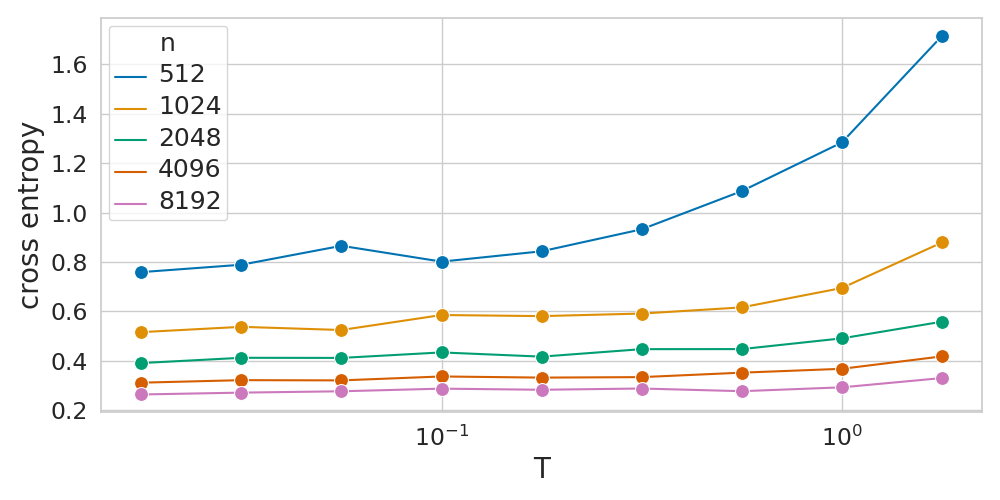}
\end{subfigure}
\begin{subfigure}[b]{0.32\textwidth}
\includegraphics[width=\textwidth]{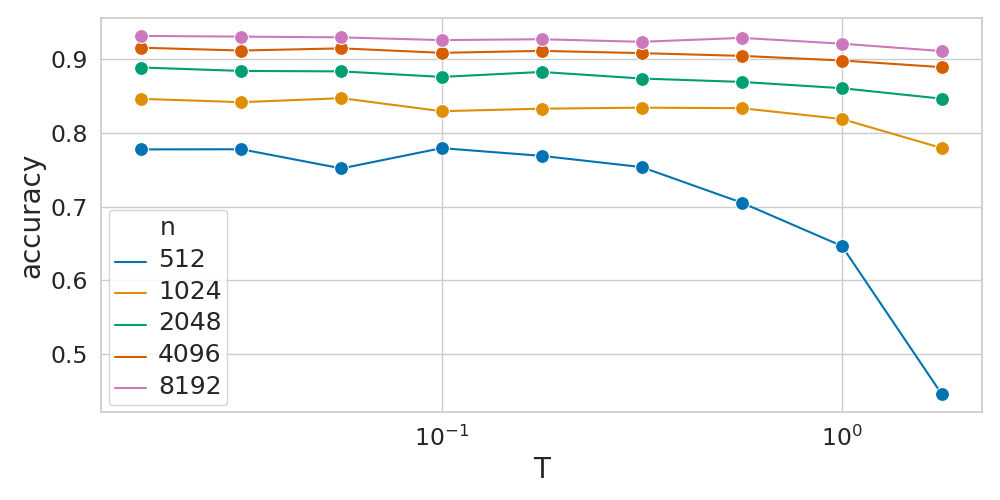}
\end{subfigure}
\begin{subfigure}[b]{0.32\textwidth}
\includegraphics[width=\textwidth]{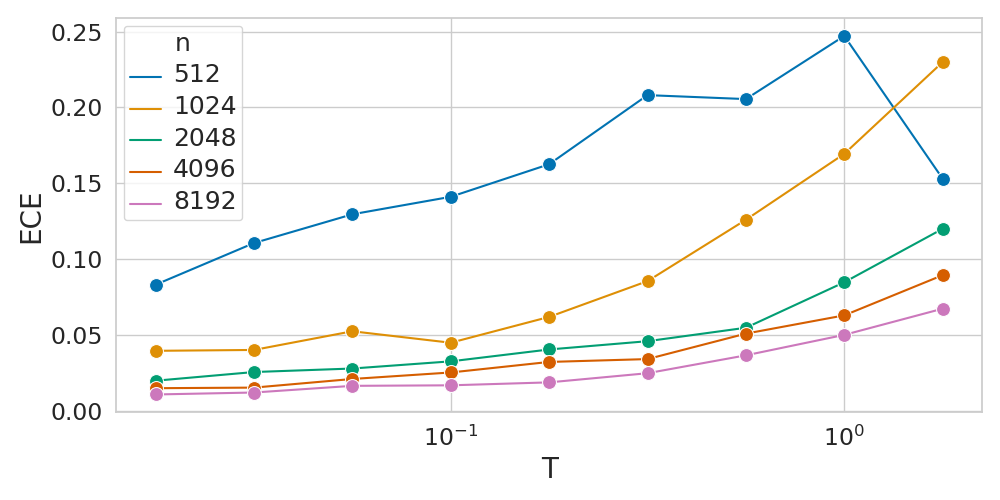}
\end{subfigure}
\caption{SVHN subsampling experiment with decreasing number of gradients steps. }
\label{fig:svhn_small_subsample}
\end{figure}

We repeat the main subsampling experiment on SVHN of Section \ref{sec:experiments}. This time we use a decreasing number of gradient steps. Results are in Fig. \ref{fig:svhn_small_subsample}. In particular, the burn-in period is always 200 epochs. The cycle length is increased from 60 at $n=8192$ to 100 at $n=512$, and total number of epochs from 1100 to 1700. Therefore the number of gradient steps decreases, as both the cycle length and the total number of training epochs should be doubled every time that the dataset is halved. 

\subsection{Experiments on CIFAR-10H}
\label{sec:cifar10h_exp}

\begin{tcolorbox}[colback=red!5!white,colframe=red!75!black]
We repeat some of the experiment on CIFAR-10H by \cite{aitchison2020statistical}. Our results show that over-weighting the likelihood with respect to the prior is sufficient for the removal of the CPE. Furthermore, this weight does not equal the number of labellers, suggesting that the curation theory of \cite{aitchison2020statistical} is not accurate enough to explain the CPE~alone (as was already acknowledged by the author).
\end{tcolorbox}

In \cite{aitchison2020statistical}, the authors state that the cold posterior effect might be due to the labelling process, based on consensus: each image is added to the dataset only if all $S$ labellers agree on one class. In particular, they say that if we had access to the $S$ original labels for each image, then we should not observe any cold posterior effect. They devise an experiment on CIFAR-10H \citep{peterson2019human}, in which all the (approximately 50) labels are given for each image. Our experiments show that over-weighting the likelihood with respect to the prior is sufficient to eliminate the CPE, and that the weight does not correspond to the number of labellers.

More formally, the dataset can be defined as $\mathcal{D} := \{ (\bm{x}_i, \bm{y}_i ) \}_{i=1}^n$, where $\bm{y}_i$ is a vector containing the \emph{counts for each class}, i.e, its $j$-th element is the number of labellers that chose class $j$ 
Given a mini-batch of size $B$, the log-likelihood used in an SG-MCMC method, is the following:
\begin{equation}
    \mathcal{L}_c := \frac{n}{B} \sum_{i=1}^B\bm{y}_i^T \log f_i .
\end{equation}
This is very similar to the use of label smoothing in which the label vector is $\bm{y}_{ls} := \frac{1}{S}\bm{y}$. The corresponding mini-batch likelihood has the form:
\begin{equation}
    \mathcal{L}_{ls} := \frac{n}{B} \sum_{i=1}^B(\bm{y}_{ls})_i^T \log f_i = \frac{n}{BS} \sum_{i=1}^B\bm{y}_i^T \log f_i .
\end{equation}
Therefore the only difference between $\mathcal{L}_c$ and the "human-aware" label-smoothing loss $\mathcal{L}_{ls}$ is that the former is $S$ times stronger, and they are conceptually very similar.  
We will also consider "standard" label smoothing with parameter $\alpha \in (0,1)$, :
\begin{equation}
    \mathcal{L}_{\alpha} := \frac{n}{B} \sum_{i=1}^B\hat{\bm{y}}_i^T \log f_i ,
\end{equation}
where $\hat{\bm{y}}_i$ is $1-\alpha$ in the position corresponding to the correct label and $\frac{\alpha}{C}$ otherwise. 

For each of the likelihoods proposed above, we train a ResNet-20 on CIFAR-10H and evaluate on CIFAR-10 training set, as in \cite{aitchison2020statistical}. Regarding the SG-MCMC method, we use cyclical step size, adapting the code from \cite{wenzel2020good}. We leave unchanged all the hyperparameters. The only exception is in the case we are using $\mathcal{L}_{c}$, where we reduce the learning rate by a factor $50$, the approximate number of labellers per image, as in \cite{aitchison2020statistical}. The reason for this reduction is that the likelihood gets $\approx 50$ times stronger due to the fact that each datapoint is labelled by $\approx 50$ labellers.  We use 100 epochs as burn-in period and a cycle length of 50 epochs for a total of 1000 epochs. We use data augmentation. \footnote{In \cite{aitchison2020statistical} it is not explicitly stated that data augmentation have been used. However, we checked the code that they use \citep{zhang2019cyclicalsgmcmc} and verified that data augmentation is indeed used there.} 

\begin{figure}[h!]
\centering
\begin{subfigure}[b]{0.4\textwidth}
\includegraphics[width=\textwidth]{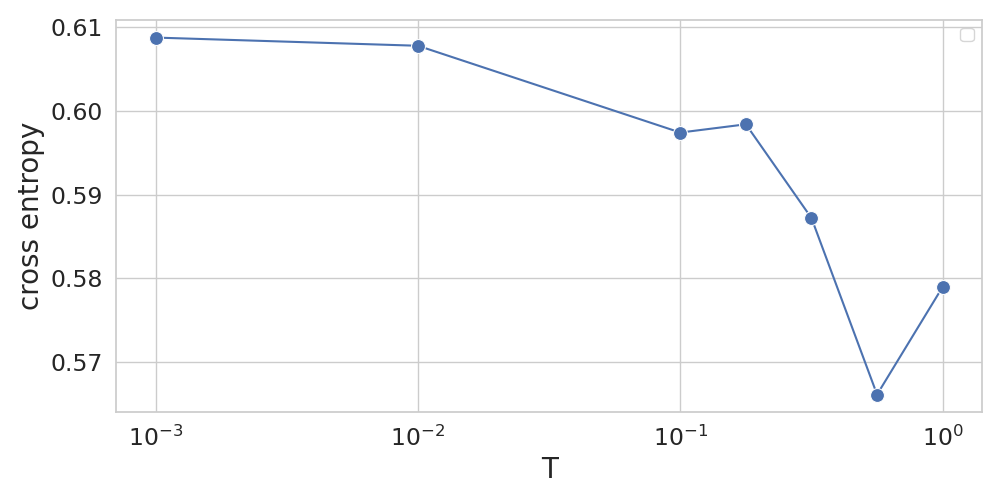}
\end{subfigure}
\begin{subfigure}[b]{0.4\textwidth}
\includegraphics[width=\textwidth]{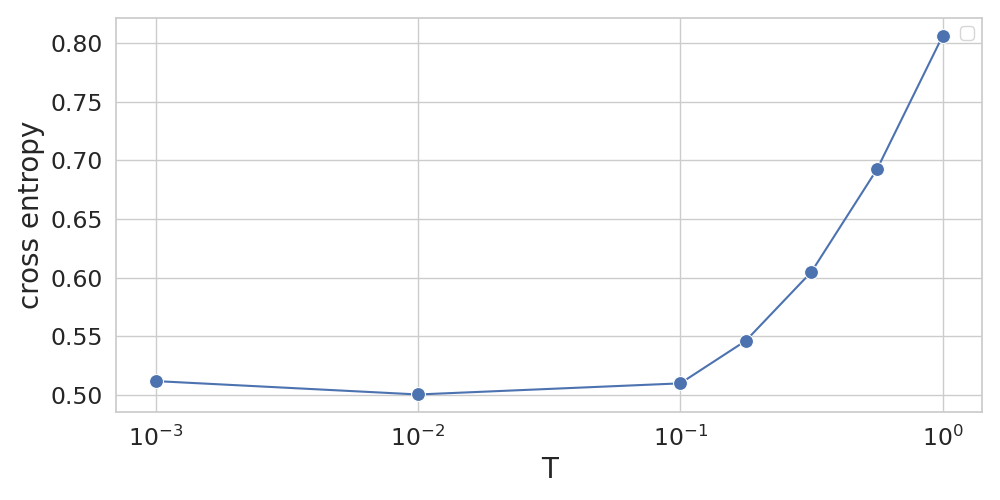}
\end{subfigure}
\caption{\textbf{left}: model with over-weighted likelihood with smoothed labels. \textbf{right}: model trained with standard label smoothing}
\label{fig:cifar10h_ls_01}
\end{figure}

\paragraph{Is it all about over-weighting the likelihood?}
We apply standard label smoothing (loss $\mathcal{L}_\alpha$) with $\alpha=0.1$ to the one hot encoded labels. Then, we overweight the likelihood by a factor of $50$ and reduce the learning by the same factor (i.e. the approximate number of labellers). Results are in Figure~\ref{fig:cifar10h_ls_01}.

Note that making the likelihood $S$ times stronger helps to eliminate the cold posterior effect. Note also that label smoothing plus likelihood over-weighting is equivalent to assuming that the labellers have assigned different labels. 

\begin{figure}[h!]
\centering
\begin{subfigure}[b]{0.4\textwidth}
\includegraphics[width=\textwidth]{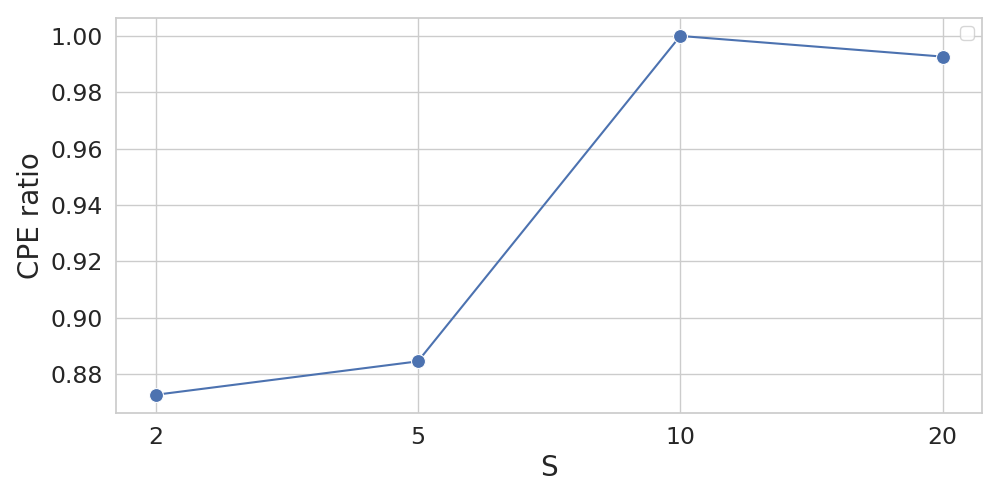}
\subcaption{}
\end{subfigure}
\begin{subfigure}[b]{0.4\textwidth}
\includegraphics[width=\textwidth]{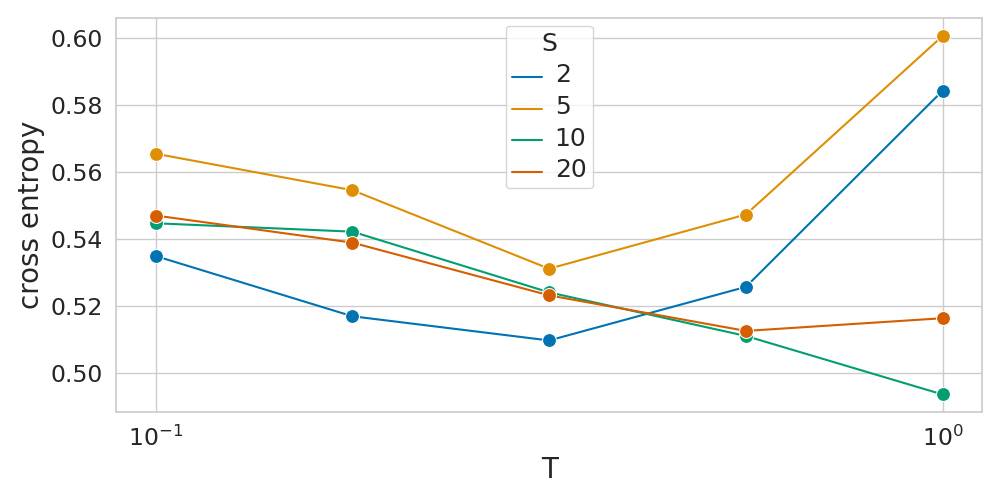}
\subcaption{}
\end{subfigure}
\caption{CPER and test CE for subsample of $\tilde{S}$ labellers from Cifar10H out of a total of $S \approx 50$ available labellers.}
\label{fig:cifar10h_subsample}
\end{figure}

\paragraph{A smaller number of labellers is enough to alleviate the cold posterior effect}
The previous experiments have shown that over-weighting the likelihood alleviates the cold posterior effect. Here, we test whether weighting the likelihood by a smaller amount is sufficient to alleviate the cold posterior effect. In particular, we subsample a different number of $\tilde{S}$ labels for each image from the counts that are available in CIFAR-10H. We do this by considering the probabilities implied by the counts, and sampling $\tilde{S}$ times from this categorical distribution, for each image.  Results are shown in Figure \ref{fig:cifar10h_subsample}. Note that $T \geq 1$ is optimal for any value of $\tilde{S} \geq 10$ that we consider, and the cold posterior effect appears only for $\tilde{S} \leq 5$. This is strong evidence that there is not an exact correspondence between the number of labellers and the optimal temperature. What matters is to over-weight the likelihood, but even a significantly smaller weight can fix the cold posterior problem.

\end{document}